\documentclass[11pt]{article}
\usepackage[preprint]{acl}
\usepackage{times}
\usepackage{latexsym}
\usepackage[T1]{fontenc}
\usepackage[utf8]{inputenc}
\usepackage{microtype}
\usepackage{inconsolata}
\usepackage{graphicx}
\usepackage{amsmath,amssymb}
\usepackage{booktabs,multirow,makecell,array,tabularx}
\usepackage{xcolor,textcomp,listings}
\usepackage{xurl}
\usepackage{xr-hyper}
\usepackage{placeins}
\usepackage{booktabs}
\usepackage{tabularx}
\definecolor{tablepink}{HTML}{F7ECEF}
\usepackage[table]{xcolor}
\usepackage{listings}
\usepackage{xcolor}
\usepackage{enumitem}
\usepackage{float}
\usepackage{amssymb}

\definecolor{MyDarkBlue}{rgb}{0,0.5,1}
\definecolor{MyDarkGreen}{rgb}{0.02,0.6,0.02}
\definecolor{MyDarkRed}{rgb}{0.8,0.02,0.02}
\definecolor{MyDarkOrange}{rgb}{0.40,0.2,0.02}
\definecolor{MyPurple}{RGB}{111,0,255}
\definecolor{MyRed}{rgb}{1.0,0.0,0.0}
\definecolor{MyGold}{rgb}{0.75,0.6,0.12}
\definecolor{MyDarkgray}{rgb}{0.66, 0.66, 0.66}

\graphicspath{{figures/}}
\newcommand{\labopcorrespondencenote}{%
  \begingroup
  \renewcommand{\thefootnote}{\fnsymbol{footnote}}%
  \footnotetext[2]{Corresponding authors.}%
  \endgroup
}
\author{%
  \begin{minipage}{\textwidth}
  \centering\normalfont\large
  \textbf{Jingyuan Li}\textsuperscript{1,2},
  \textbf{Yujie Liu}\textsuperscript{1,3},
  \textbf{Yijie Dai}\textsuperscript{2},
  \textbf{Xiaoyu Yuan}\textsuperscript{4},
  \textbf{Nian Ran}\textsuperscript{5},\\[1pt]
  \textbf{Huajie Niu}\textsuperscript{6},
  \textbf{Ming Zhu}\textsuperscript{5},
  \textbf{Shuwen Zhao}\textsuperscript{5},
  \textbf{Chenxi Li}\textsuperscript{1,7},
  \textbf{Fenghao Lu}\textsuperscript{8},\\[1pt]
  \textbf{Jiahao Li}\textsuperscript{8},
  \textbf{Xinyu Gu}\textsuperscript{1},
  \textbf{Zhe Liu}\textsuperscript{9},
  \textbf{Long Xu}\textsuperscript{2},
  \textbf{Wenlong Zhang}\textsuperscript{1},
  \textbf{Bo Zhang}\textsuperscript{1},\\[1pt]
  \textbf{Lei Bai}\textsuperscript{1},
  \textbf{Peiyu Zhang}\textsuperscript{4\textdagger},
  \textbf{Minting Pan}\textsuperscript{1\textdagger},
  \textbf{Weinan Zhang}\textsuperscript{1,2\textdagger},
  \textbf{Dongzhan Zhou}\textsuperscript{1\textdagger}\\[3pt]
  {\normalfont\normalsize
  \textsuperscript{1} Shanghai Artificial Intelligence Laboratory\quad
  \textsuperscript{2} Shanghai Jiao Tong University\\[1pt]
  \textsuperscript{3} Fudan University\quad
  \textsuperscript{4} XtalPi\quad
  \textsuperscript{5} Shanghai Institute of Ceramics, Chinese Academy of Sciences\\[1pt]
  \textsuperscript{6} Jiaxing University\quad
  \textsuperscript{7} Zhejiang University\quad
  \textsuperscript{8} University of Chinese Academy of Sciences\\[1pt]
  \textsuperscript{9} East China University of Science and Technology\\[5pt]
  \texttt{johnnylee@sjtu.edu.cn, peiyu.zhang@xtalpi.com}\\[1pt]
  \texttt{\{panminting,zhangweinan,zhoudongzhan\}@pjlab.org.cn}
  }
  \end{minipage}%
}

\title{LabOPBench: A Multimodal Benchmark for Evidence-Grounded Laboratory Decision Making}
\begin{document}
\maketitle
\labopcorrespondencenote
\begin{abstract}
Laboratory experiments involve more than following predefined protocols; they require adaptive decision-making as the experimental state changes. Existing scientific benchmarks rarely assess such evidence-based decisions in laboratory settings. 
Here we present LabOPBench, a multimodal benchmark of $756$ expert-authored, independently reviewed questions in organic synthesis and materials science. Each item pairs laboratory evidence with procedural context and an open-ended judgment, evaluated using a question-specific scoring rubric. 
Eight multimodal models obtain overall normalized scores of $28.8$-$45.8\%$, leaving substantial room for improvement.
On a controlled $100$-question subset, the strongest models outperform undergraduate participants on average but remain below the average performance of PhD participants. 
Expert analysis of responses from the three leading models identifies Evidence Misreading and Evidence-Decision Conflict as the dominant error patterns.
These results reveal that current models still struggle to turn multimodal perception into reliable laboratory decision-making.


\end{abstract}

\section{Introduction}
\label{sec:introduction}

Large language models (LLMs) and LLM-based agents increasingly support laboratory research through experimental planning, protocol generation, and integration with laboratory instruments and robotic systems~\cite{boiko2023autonomous,darvish2025organa,hase2019sdl,tom2024sdl,coley2019flow,burger2020mobile}. Yet laboratory research involves more than executing predefined procedures. As experiments progress, unexpected observations and changes in sample states may require researchers to verify, adjust, or terminate an operation. Evaluating multimodal models for laboratory use requires assessing their ability both to generate scientifically plausible procedures and to make appropriate judgments as experiments evolve.

Laboratory observations cannot be interpreted in isolation. The same visual pattern may indicate different causes and require different actions depending on the experimental context. Although previous studies have highlighted challenges in visual grounding and multimodal chemical reasoning~\cite{tong2024eyeswide,cui2025chemolympiad}, recognizing an observation is only the first step. A model should also determine whether its explanation or proposed action is appropriate for the specific experiment. If the evidence is insufficient, further checks may be needed before drawing conclusions or taking action. Evidence-to-decision consistency describes how well laboratory judgments align with experimental evidence and context.


Existing scientific benchmarks evaluate a broad range of capabilities, including domain knowledge, multimodal reasoning, laboratory safety, and experimental procedures~\cite{laurent2024labbench,alampara2025macbench,zhou2026labsafety,zhang-etal-2026-chemreason}. However, these evaluations provide less insight into whether models can use experimental observations to make appropriate laboratory judgments. In practice, recognizing an observation is only part of the task. Models must also consider the experimental context to determine what the observation means and how to respond. This motivates evaluating whether a model can make laboratory judgments supported by the available evidence and experimental context.


We introduce \textbf{LabOPBench}, a multimodal benchmark of $756$ expert-authored and independently reviewed questions in organic synthesis and materials science. Constructing such a benchmark requires substantial scientific judgment, as real laboratory materials do not readily translate into well-defined, objectively assessable decision tasks. To address this challenge, $22$ contributors with hands-on laboratory experience developed open-ended questions from real laboratory materials. For each question, they identified decision-relevant evidence, provided the necessary experimental context, and defined judgments supported by the available information. They also considered alternative explanations or actions, ensuring that each question could be answered without relying on missing information or unresolved ambiguity. Experts then developed question-specific scoring rubrics, while two additional reviewers independently assessed each item for scientific validity, evidence sufficiency, and scoring clarity.

Evaluation of eight multimodal models on LabOPBench reveals substantial limitations in evidence-grounded laboratory decision-making. Overall normalized scores range from $28.8\%$ to $45.8\%$, indicating considerable room for improvement. Performance varies across task categories, with postprocessing showing the lowest cross-model mean ($34.9\%$), compared with materials science ($52.2\%$) and monitoring ($50.8\%$). Postprocessing tasks present particular challenges, requiring integration of preceding experimental operations and current sample states. Expert analysis further identifies recurring failures to interpret decision-relevant observations or align explanations and actions with the experimental context. Together, these findings suggest that current models still face challenges in interpreting laboratory evidence and using it to make reliable decisions in specific experimental situations.

Overall, the contributions of this paper are:

\begin{itemize}
\vspace{-3pt}
\item We introduce LabOPBench, a multimodal benchmark of $756$ expert-authored questions grounded in real laboratory materials from organic synthesis and materials science. It evaluates how models interpret experimental states, identify possible causes and risks, and determine appropriate operations.
\vspace{-3pt}
\item We establish expert-defined, question-specific scoring rubrics to assess whether laboratory judgments are supported by the available evidence and experimental context. These rubrics assess whether answers are supported by the evidence and fit the specific experimental context, rather than merely being scientifically plausible. 
\vspace{-3pt}
\item We systematically evaluate eight multimodal models
to characterize their performance and limitations in
laboratory decision-making. Our findings show that models often misinterpret experimental evidence or make judgments that do not fit the laboratory context.
\end{itemize}




\section{Related Work}
\label{sec:related}

\paragraph{Scientific knowledge benchmarks.}
Existing chemistry benchmarks evaluate molecular and reaction reasoning, factual chemical knowledge, quantitative problem solving, and expert-level question answering~\cite{guo2023chemllmbench,lu2024moleculeqa,huang2026chemeval,mirza2025chembench,runcie2026chemiq,xie2025qcbench}. Recent multimodal benchmarks further incorporate molecular structures, reaction schemes, chemical tables, characterization images, and scientific documents~\cite{li2025chemvlm,zhou2025chemtable,lai2025matcha,li2025rxnbench,alampara2025macbench}. More general multimodal evaluations, such as MMMU-Pro, have also emphasized the importance of genuine visual understanding and cross-modal reasoning~\cite{yue2025mmmupro}. Together, these resources provide important assessments of scientific knowledge, visual interpretation, and multimodal reasoning.

\paragraph{Laboratory procedural benchmarks.}
Recent benchmarks have increasingly addressed practical scientific research tasks. LAB-Bench evaluates biological research capabilities, including protocol troubleshooting, figure interpretation, and experimental planning~\cite{laurent2024labbench}. MaCBench examines multimodal chemistry and materials tasks spanning data extraction, experimental execution, and results interpretation~\cite{alampara2025macbench}, while MatCha focuses on expert-level understanding of materials characterization images~\cite{lai2025matcha}. LabSafety Bench evaluates laboratory hazard identification, risk assessment, and consequence prediction using multiple-choice questions and open-ended scenarios~\cite{zhou2026labsafety}. 
ChemReason-Bench~\cite{zhang-etal-2026-chemreason} evaluates verifiable procedural reasoning in organic synthesis through ordering, validation, constrained completion, selection, and explanation tasks. While these benchmarks assess practical laboratory skills and procedural consistency, LabOPBench focuses on open-ended laboratory judgments grounded in multimodal
evidence and experimental context.
Table~\ref{tab:benchmark_comparison}
summarizes the key differences.

\begin{table*}[t]
\centering
\small
\setlength{\tabcolsep}{5pt}
\renewcommand{\arraystretch}{1.20}

\begin{tabularx}{\textwidth}{@{}
>{\raggedright\arraybackslash}p{0.35\textwidth}
>{\raggedright\arraybackslash}p{0.18\textwidth}
>{\centering\arraybackslash}p{0.15\textwidth}
>{\raggedright\arraybackslash}X
@{}}

\toprule
\textbf{Benchmark} &
\textbf{Data source} &
\textbf{Format} &
\textbf{Focus} \\
\midrule

LAB-Bench~\cite{laurent2024labbench} &
Literature / protocols &
MCQ &
Biological research \\

MMMU-Pro~\cite{yue2025mmmupro} &
Academic exams &
MCQ &
Visual reasoning \\

MatCha~\cite{lai2025matcha} &
Published figures &
MCQ &
Materials characterization \\

MaCBench~\cite{alampara2025macbench} &
Figures / lab photos &
MCQ + Numeric &
Data \& experiment analysis \\

LabSafety Bench~\cite{zhou2026labsafety} &
Safety resources &
MCQ + Open &
Hazards \& consequences \\

ChemReason-Bench~\cite{zhang-etal-2026-chemreason} &
Synthesis protocols &
Mixed &
Procedural reasoning \\

\midrule

\textbf{LabOPBench} &
\textbf{Original lab records} &
\textbf{Open} &
\textbf{Evidence-based decisions} \\

\bottomrule
\end{tabularx}

\caption{Comparison of LabOPBench with related benchmarks by data source, format, and focus.}
\label{tab:benchmark_comparison}
\end{table*}

\paragraph{Evidence-Grounded Laboratory Reasoning.}
Laboratory perception studies have investigated the recognition of vessels, material phases, liquid levels, phase boundaries, turbidity, precipitates, and color changes in experimental settings~\cite{eppel2020labpics,elkhawaldeh2024eye,elkhawaldeh2024lle}. Related research has also explored multimodal analysis of laboratory anomalies and experimental states~\cite{lin2025sdldataset,lin2025vlmanomaly}. These studies demonstrate the importance of extracting and interpreting visual evidence from laboratory environments. However, identifying an experimental phenomenon or abnormal state does not guarantee an appropriate explanation or action. Such judgments must also be consistent with the available evidence and previous experimental steps. LabOPBench focuses on the evidence-to-decision relationship by evaluating open-ended judgments about experimental states, possible causes, and appropriate actions. Through expert-defined reference judgments and question-specific scoring rubrics, it assesses whether models connect multimodal observations and procedural context to experimentally appropriate conclusions.

\section{LabOPBench}
\label{sec:method}

\subsection{Task Definition}
\label{sec:task_definition}

LabOPBench evaluates whether a multimodal model can transform evidence from a concrete laboratory situation into an experimentally appropriate judgment. Each benchmark item contains an experimental context, one or more evidence sources, and an open-ended question. The evidence may include laboratory photographs, instrument interfaces, spectra, chromatograms, microscopy images, sample states, or experimental records. Given these inputs, a model is expected to identify the relevant experimental state, explain a likely cause or risk when required, and select an appropriate next action.
Formally, each item can be represented as
\[
q_i=(c_i,e_i,p_i,r_i),  
\]
where $c_i$ denotes the experimental context, $e_i$ the associated multimodal evidence, $p_i$ the question prompt, and $r_i$ the question-specific scoring rubric. The answer model observes only $(c_i,e_i,p_i)$. The target judgment may concern the current state of an experiment, a likely cause of an observation, a potential risk, or an appropriate experimental operation. The task therefore requires evidence selection and contextual interpretation rather than isolated object recognition or factual question answering.

\subsection{Benchmark Composition}
\label{sec:composition}

\begin{figure*}[t]
\centering
\includegraphics[width=\linewidth]{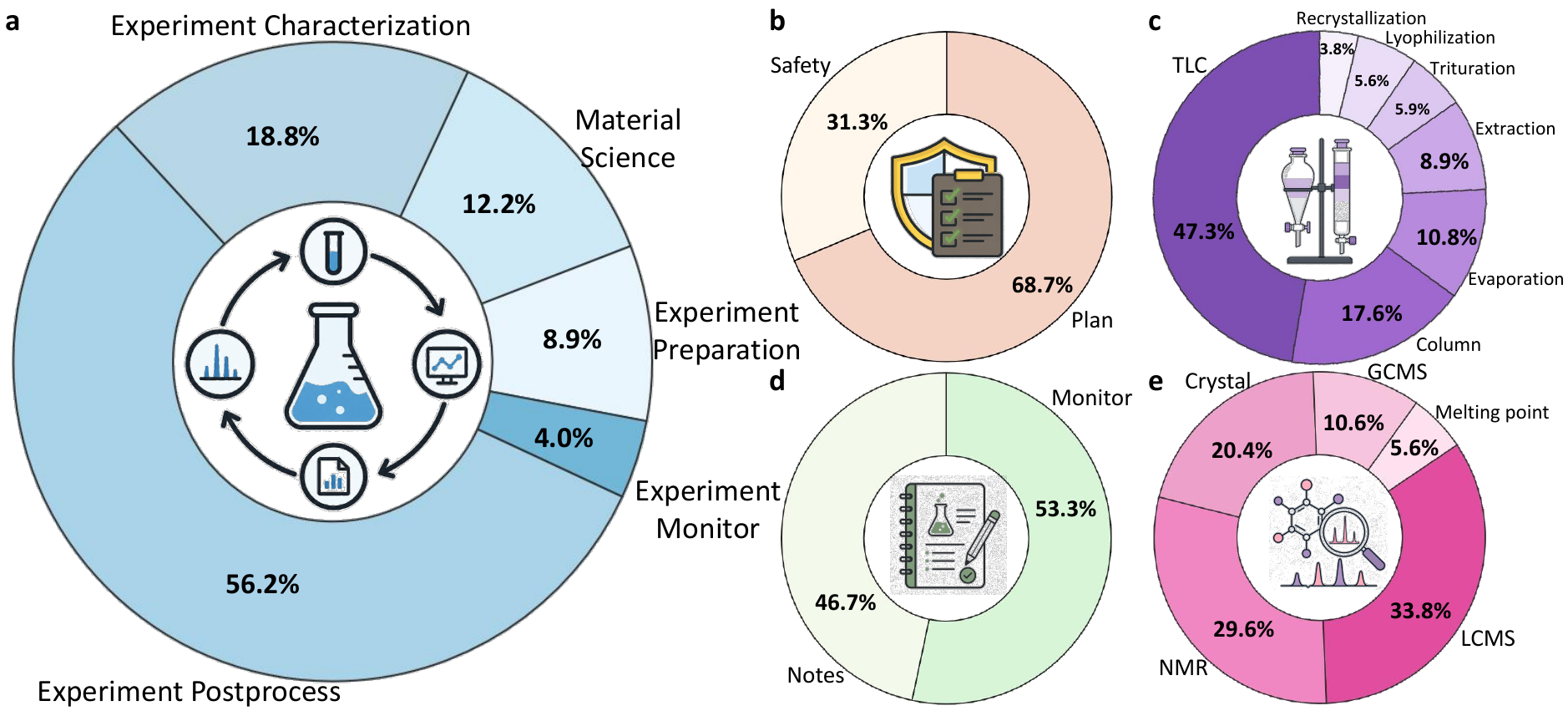}
\caption{Composition of LabOPBench across laboratory task categories. \textbf{a}, Distribution of the 756 benchmark questions across the five major categories. \textbf{b-e}, Distribution of organic-synthesis questions across the subcategories of preparation, postprocessing, monitoring and characterization, respectively.}
\label{fig:question_all}
\end{figure*}

LabOPBench contains $756$ questions, including $664$ questions from organic synthesis and 92 questions from materials science (Fig.~\ref{fig:question_all}). The organic-synthesis questions are organized according to the stage of the experimental workflow in which the target judgment arises: preparation ($67$), monitoring ($30$), postprocessing ($425$), and characterization ($142$). Materials-science questions are retained as a separate major category because their experimental objects, procedures, and measurements do not map directly onto the organic-synthesis taxonomy. The category sizes reflect the assessable situations available in the collected laboratory materials rather than predefined quotas. Each judgment must be grounded in the supplied evidence, experimental context, and scientific knowledge.

\subsection{Question Authoring and Expert Review}
\label{sec:question_authoring}

LabOPBench questions were authored by $22$ contributors with hands-on experience relevant to the represented laboratory tasks. The contributor pool comprised four PhD researchers, six postdoctoral researchers, eight laboratory specialists, and four researchers at associate-researcher level or above. Twenty contributors had at least $5$ years of laboratory experience. Their expertise spanned laboratory workflows, from experimental setup to characterization and troubleshooting. Further details are provided in Appendix~\ref{app:contributors}.

Each item was designed from real laboratory materials to assess a well-defined experimental judgment. We view expert authoring as a transformation of original laboratory material into a structured decision instance:
\[
\Phi_{\mathrm{expert}}(M_i)
\longrightarrow
\mathcal{B}_i
=
\left(
C_i,E_i,Q_i,Y_i,\Pi_i,\mathcal{R}_i
\right),
\]
where $M_i$ denotes the original laboratory material; $C_i$, $E_i$, $Q_i$, and $Y_i$ denote the experimental context, supplied evidence, question prompt, and target experimental judgment, respectively. $\Pi_i$ denotes the expert-specified evidence-to-decision path, and $\mathcal{R}_i$ denotes the associated scoring rubric. Depending on the item, $\Pi_i$ connects evidence to an experimental state or an appropriate action.

The scientific content of every item was authored and validated by human experts. Language models were not used to generate the experimental scenario, target judgment, evidence selection, reference answer, acceptable alternatives, or scoring criteria. For each item, contributors recorded the decision-relevant evidence, acceptable alternative explanations or actions, critical errors, and four scoring anchors corresponding to scores of $3$, $2$, $1$, and $0$. Questions were required to be answerable from the supplied evidence and relevant scientific knowledge without relying on undisclosed experimental conditions or author-specific information.

DeepSeek-V4-Pro was used only for post hoc linguistic standardization of question prompts and reference answers after the scientific content had been fixed. It was not permitted to change the target judgment, decision-relevant evidence, acceptable answers, scoring anchors, or safety requirements. The original author verified each standardized version before two additional experts independently reviewed the credibility and sufficiency of the evidence, prompt-evidence consistency, scientific correctness, answer determinacy, and scoreability. Substantive disagreements were adjudicated by a third expert, whereas questions with insufficient evidence or unresolved ambiguity were revised or excluded. The question-construction and expert-validation workflow is summarized in Fig.~\ref{fig:workflow}. Further details are provided in Appendix~\ref{app:question-construction}.

\begin{figure*}[t]
\centering
\includegraphics[width=\linewidth]{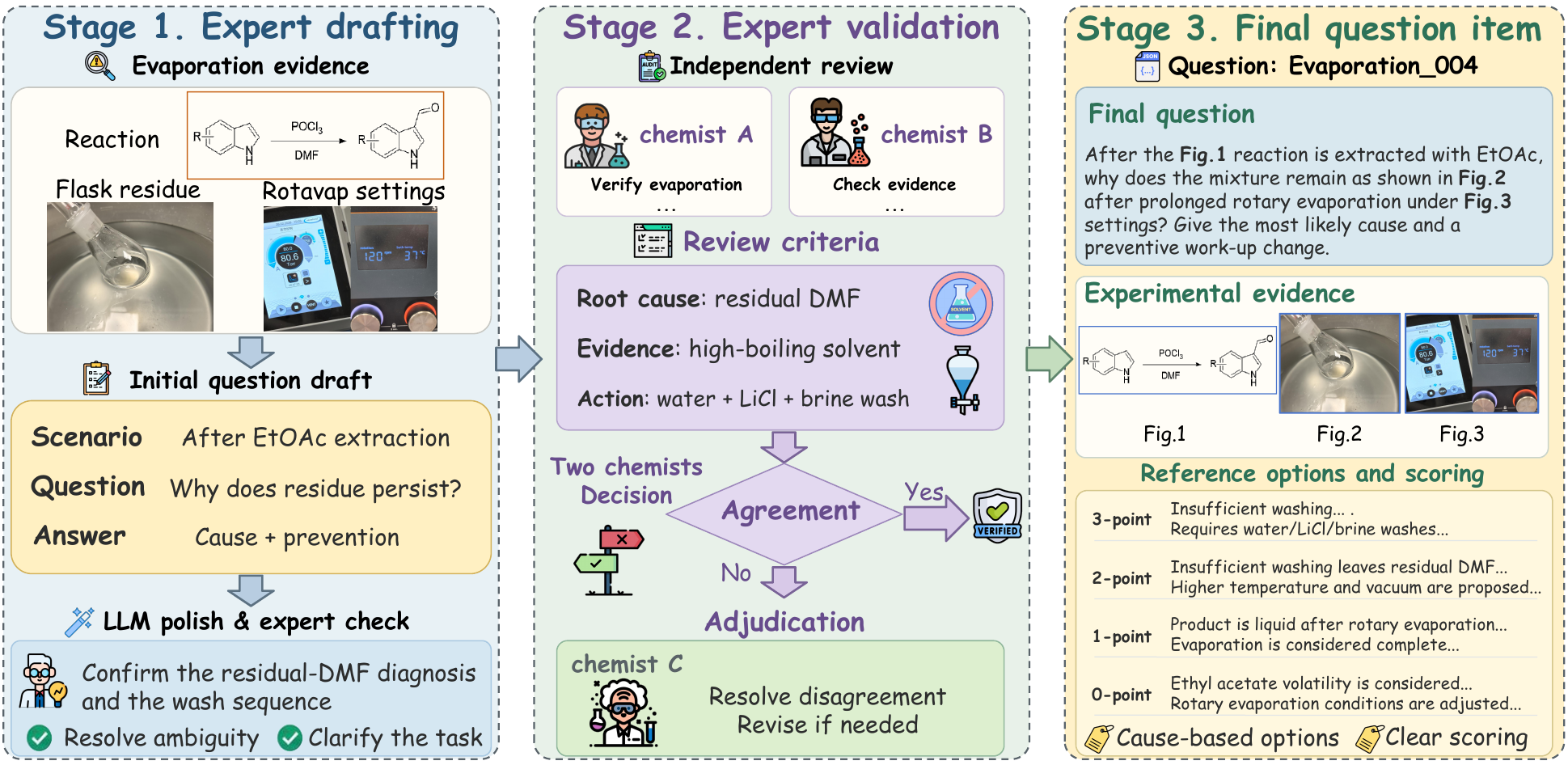}
\caption{Workflow for constructing and validating a LabOPBench question. Experts draft and review each evidence-grounded item, while an LLM is used only for language standardization. Disagreements are adjudicated before the final question, multimodal evidence, reference options, and scoring anchors are compiled.}
\label{fig:workflow}
\end{figure*}

\subsection{Evaluation Protocol}
\label{sec:evaluation_protocol}

We evaluated eight multimodal models on all $756$ questions
under a unified input and generation protocol. Each model
received the question, its minimal experimental context,
and the original visual evidence, without access to reference
answers, scoring rubrics, or external tools.

Responses were evaluated with question-specific scoring rubrics on a $0$-$3$ scale, with half-point scores permitted. All eight evaluated models produced valid responses for all $756$ benchmark questions. GPT-5.5 and Gemini 3.1 Pro each assigned a valid score to every response from every evaluated model; therefore, \(N_j=756\) for each model-judge pair when reporting the full benchmark. The Overall score in Table~\ref{tab:full-benchmark} is calculated across all $756$ benchmark questions. A higher score required the response to satisfy the scenario-specific evidence, interpretation, and action requirements of the corresponding rubric level; general plausibility or keyword overlap was insufficient. Holistic scoring was used by default, while explicitly independent subquestions were scored separately and averaged. GPT-5.5 and Gemini 3.1 Pro independently scored each response using the question, candidate response, and reference answer. The judges were blinded to the evaluated model and to each other's assessments. For a reporting set containing \(N_j\) valid responses, the normalized score assigned by judge \(j\) was
\[
\mathrm{NS}_j=
\frac{1}{3N_j}
\sum_{i=1}^{N_j}s_{ij}\times 100\%,
\]
where $s_{ij}$ is the score assigned to question $i$. The final model score was the arithmetic mean of the two judge-specific scores. Detailed scoring and aggregation procedures are provided in Appendix~\ref{app:score}.

We validated automated scoring using a stratified sample of approximately $20$\% of Claude Opus 4.7 responses from each major category. Two experts independently rescored the sampled responses with the same rubrics, without access to automated scores or each other's ratings, and agreement was assessed at the category-mean level. 

We further used a controlled $100$-question organic-synthesis subset for comparison with five undergraduate and five PhD participants, an image-mismatch condition for testing image-question alignment, and expert annotation of errors in responses from the three highest-scoring models. These analyses used the same scoring protocol; detailed procedures are provided in Appendix~\ref{app:100-question subset}.

\section{Experiments}
\label{sec:experiments}
We evaluated eight multimodal models: Claude Opus 4.7~\cite{anthropic2026claudeopus47}, GPT-5.5~\cite{openai2026gpt55}, Gemini 3.1 Pro~\cite{google2026gemini31pro}, Gemini 3.1 Flash-Lite~\cite{google2026gemini31flashlite}, Qwen3.6-Plus~\cite{qwenteam2026qwen36plus}, MiniMax-M3~\cite{minimax2026m3}, Grok 4.3~\cite{xai2026grok43}, and Intern-S1-Pro~\cite{zou2026interns1pro}. All eight models were evaluated on the full set of $756$
LabOPBench questions using a unified protocol.
Each model received the question, experimental context,
and multimodal evidence, without access to reference
answers, scoring rubrics, or external tools.
GPT-5.5 and Gemini 3.1 Pro independently scored the
open-ended responses using expert-defined,
question-specific rubrics. The two normalized judge
scores were then averaged.
We further compared model and human performance, examined the effects of image mismatch, and analyzed model error patterns.
Detailed settings and procedures are provided in
Appendices~\ref{app:Evaluation}-\ref{app:error}.

\begin{table*}[t]
\centering
\small
\setlength{\tabcolsep}{3.5pt}
\renewcommand{\arraystretch}{1.00}

\begin{tabular*}{\textwidth}{@{\extracolsep{\fill}}lcccccc@{}}
\toprule
Model
& \makecell[c]{Preparation}
& \makecell[c]{Monitoring}
& \makecell[c]{Postprocessing}
& \makecell[c]{Characterization}
& \makecell[c]{Materials Science}
& \makecell[c]{\textbf{Overall}}
\\
\midrule

Intern-S1-Pro
& 24.2
& 40.0
& 25.6
& 29.6
& 41.7
& 28.8 \\

Gemini 3.1 Flash-Lite
& 41.1
& 41.4
& 33.1
& 37.8
& 47.6
& 36.8 \\

Grok 4.3
& 35.8
& 49.4
& 33.5
& 37.4
& 53.7
& 37.5 \\

MiniMax-M3
& 47.3
& \underline{54.8}
& 33.3
& 41.2
& \underline{55.6}
& 39.6 \\

Qwen3.6-Plus
& 37.3
& \underline{54.8}
& 37.2
& 44.8
& 52.8
& 41.2 \\

Gemini 3.1 Pro
& 38.0
& 48.1
& 37.5
& \textbf{48.0}
& 53.5
& 41.9 \\

GPT-5.5
& \underline{49.0}
& 52.5
& \underline{39.0}
& 47.1
& 52.9
& \underline{43.6} \\

Claude Opus 4.7
& \textbf{49.1}
& \textbf{65.0}
& \textbf{40.3}
& \underline{47.4}
& \textbf{59.8}
& \textbf{45.8} \\

\bottomrule
\end{tabular*}

\caption{Normalized scores (\%) of eight multimodal models across the major LabOPBench categories. Bold and underlined values denote the highest and second-highest scores in each column, respectively.}
\label{tab:full-benchmark}
\end{table*}

\begin{figure*}[t]
\centering
\includegraphics[width=\linewidth]{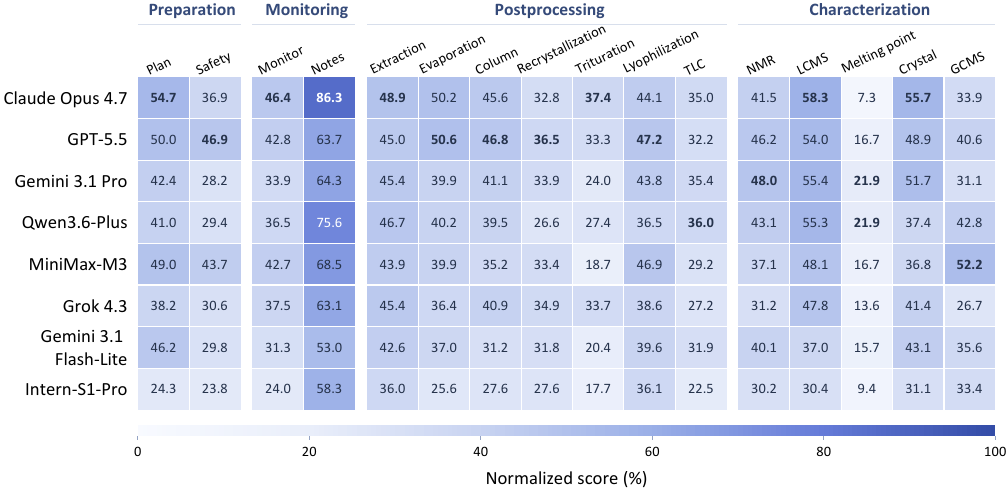}
\caption{Fine-grained performance of eight multimodal
models across 16 organic-synthesis subcategories.
Cells show the mean normalized scores (\%) from
two independent judges, with darker colors
indicating higher scores. Bold values indicate
the highest score in each subcategory, including ties.}
\label{fig:subresult}
\end{figure*}

\subsection{Main Results}
\label{sec:full_benchmark_performance}

We first evaluate the eight multimodal models on the complete LabOPBench. Evaluation settings and scoring follow the protocol described in Section~\ref{sec:evaluation_protocol}. As shown in Table~\ref{tab:full-benchmark}, normalized scores range from $28.8$\% to $45.8$\%. Claude Opus 4.7 achieves the highest overall score ($45.8$\%), followed by GPT-5.5 ($43.6$\%) and Gemini 3.1 Pro ($41.9$\%).

Performance varied substantially across the five major categories. The highest cross-model mean scores were obtained in materials science ($52.2$\%) and monitoring ($50.8$\%), followed by characterization ($41.7$\%) and preparation ($40.2$\%). Postprocessing was the most challenging category, with a cross-model mean of $34.9$\%. Notably, postprocessing was the lowest-scoring category for seven of the eight models, suggesting that its difficulty was not driven by a single model. Even Claude Opus 4.7 achieved only $40.3$\% in this category. These results suggest greater difficulty with the experimental judgments represented in postprocessing tasks, which often involve interpreting observations in relation to preceding operations and sample states. Claude Opus 4.7 achieved the highest score in four of the five categories, while Gemini 3.1 Pro achieved the highest score in characterization. Figure~\ref{fig:subresult} provides a finer-grained view of performance across the 16 organic-synthesis subcategories.

\begin{figure}[t]
\centering
\includegraphics[width=\linewidth]{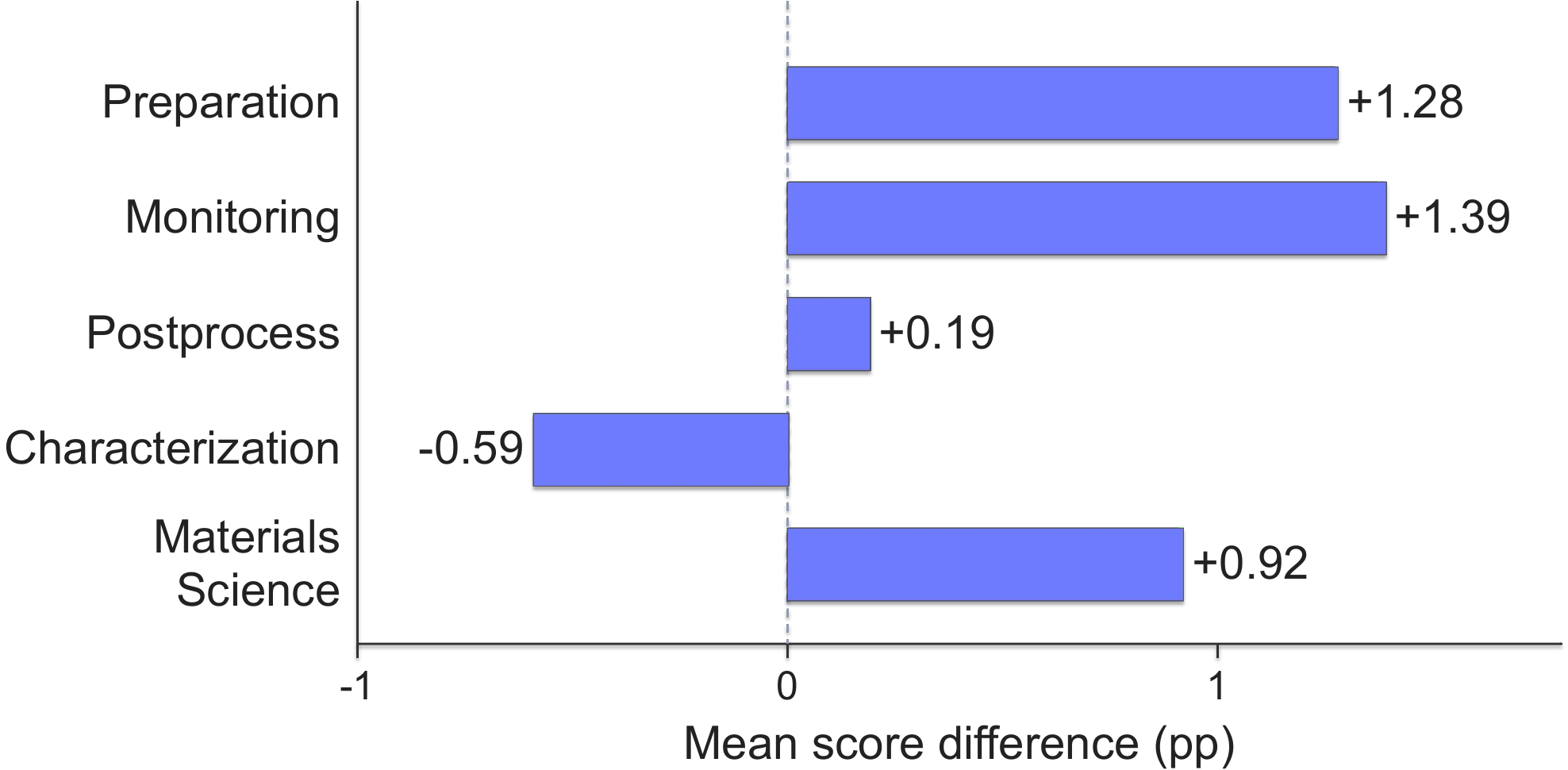}
\caption{Differences between automated and expert scoring across major laboratory categories. Bars show differences in category-level mean normalized scores (automated minus expert) for a stratified sample of Claude Opus 4.7 responses. Positive values indicate higher automated scores, whereas negative values indicate higher expert scores. }
\label{fig:Consistency}
\end{figure}

We also compared automated scores with expert assessments on Claude Opus 4.7 responses, as described in Appendix~\ref{app:Validation}. Automated and expert scores showed close agreement at the category level, with differences in mean scores across the five major categories ranging from $-0.59$ to $+1.39$ percentage points (Fig.~\ref{fig:Consistency}).

\begin{figure}[t]
\centering
\includegraphics[width=\linewidth]{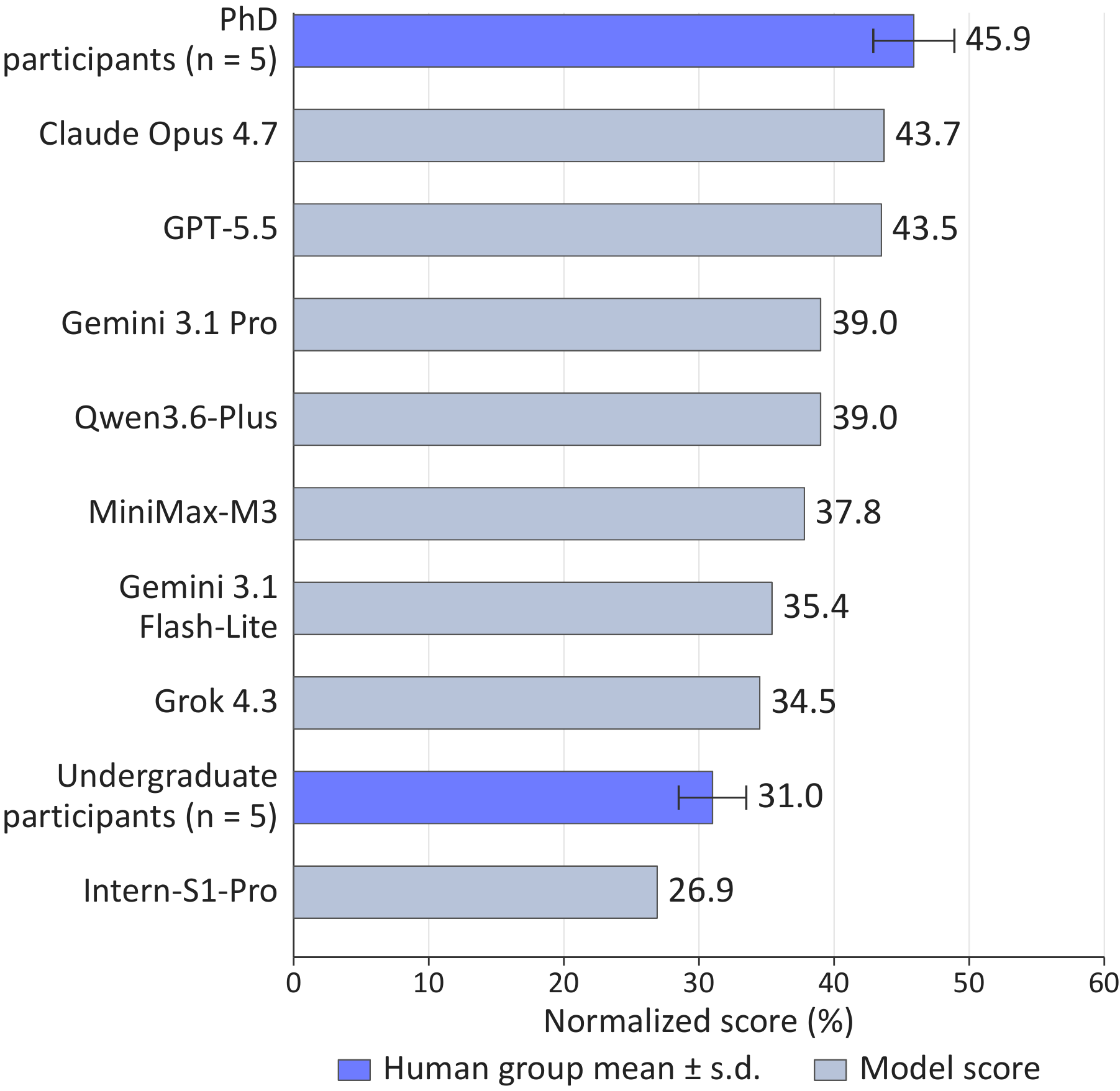}
\caption{Performance on the controlled 100-question organic-synthesis subset. Bars show normalized scores for eight multimodal models and mean normalized scores for undergraduate and PhD participants.}
\label{fig:human-answer}
\end{figure}

\subsection{Comparison with Human Participants}
\label{sec:human_results}
We compared models with human participants on the controlled
$100$-question organic-synthesis subset (Section~\ref{sec:evaluation_protocol}).
Undergraduates averaged $31.0\%$, compared with $45.9\%$ for
PhD participants (Fig.~\ref{fig:human-answer}).
Claude Opus 4.7 and GPT-5.5 were the top-performing models,
scoring $43.7\%$ and $43.5\%$.

Seven of the eight models exceeded the undergraduate-group mean, while all eight remained below the PhD-group mean. Notably, Claude Opus 4.7 was only $2.2$ percentage points below the PhD average. These results show how model performance compares with that of undergraduate and PhD participants.

\subsection{Analysis of Visual Evidence}
\label{sec:visual_results}
\begin{figure}[t]
\centering
\includegraphics[width=\linewidth]{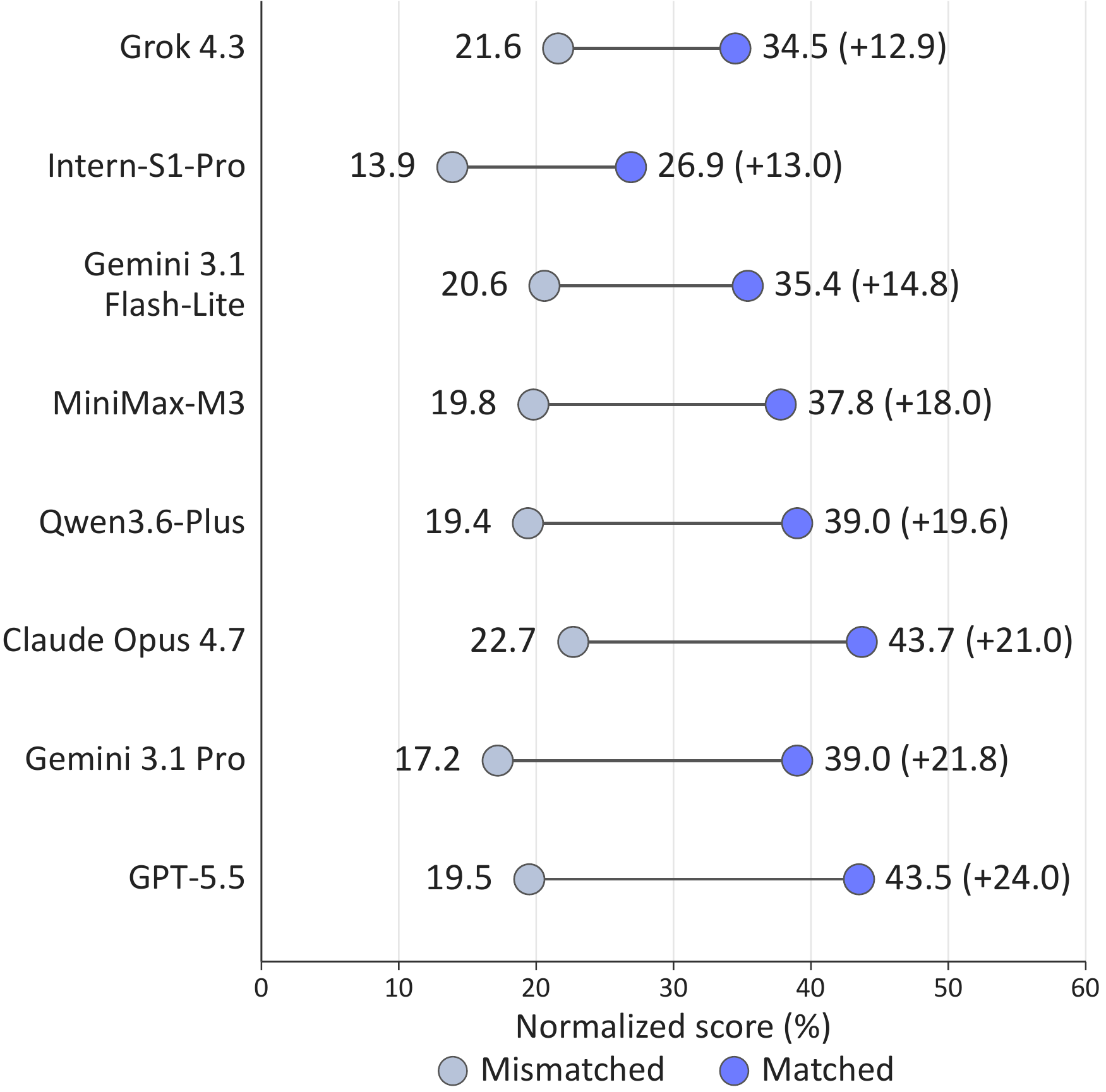}
\caption{Image-mismatch ablation on the 100-question organic-synthesis subset. Connected points show normalized scores with matched and mismatched images. Values in parentheses indicate the score difference (matched minus mismatched) in percentage points.}
\label{fig:ablation}
\end{figure}

To examine the role of question-specific visual evidence in laboratory judgments, we compared model performance under matched- and mismatched-image conditions. The experiment was conducted on the same controlled 100-question organic-synthesis subset used for human comparison. The image-mismatch procedure preserved the question format and task category while disrupting the correspondence between the images and the experimental situation (Fig.~\ref{fig:ablation}).

Image mismatch reduced performance for all eight models. The cross-model mean decreased from $37.5$\% with matched images to $19.3$\% with mismatched images, corresponding to an $18.2$-percentage-point drop. The largest decreases were observed for GPT-5.5, from $43.5$\% to $19.5$\% ($24.0$ percentage points), Gemini 3.1 Pro, from $39.0$\% to $17.2$\% ($21.8$ percentage points), and Claude Opus 4.7, from $43.7$\% to $22.7$\% ($21.0$ percentage points).

The consistent performance decline across all models indicates that their laboratory judgments were sensitive to the correspondence between visual evidence and the experimental context. However, the magnitude of this sensitivity varied across models, with score reductions ranging from $12.9$ to $24.0$ percentage points. Notably, several of the strongest models under matched-image conditions also experienced substantial decreases when this correspondence was disrupted. These findings highlight the importance of question-specific visual evidence in experimental judgment.

\begin{table*}[t]
\centering
\small
\setlength{\tabcolsep}{5pt}
\renewcommand{\arraystretch}{1.22}
\begin{tabularx}{\textwidth}{@{}
>{\raggedright\arraybackslash}p{0.08\textwidth}
>{\raggedright\arraybackslash}p{0.30\textwidth}
>{\raggedright\arraybackslash}p{0.27\textwidth}
>{\raggedright\arraybackslash}X
@{}}
\toprule
\textbf{Case} &
\textbf{Expert judgment} &
\textbf{Model response} &
\textbf{Failure mechanism} \\
\midrule
\textbf{TLC}
&
Spotting was standard; the developed
pattern indicated unsuitable
developing-solvent polarity.
&
Judged spotting as non-standard
and attributed the pattern to
excessive sample loading.
&
\emph{Evidence Misreading.}
Misinterpreted spotting evidence
and selected an unsupported cause.
\\
\addlinespace[5pt]
\textbf{LC-MS sampling}
&
Collect a representative
solid-liquid aliquot while
stirring, then dissolve
in methanol.
&
Proposed centrifugation or
filtration, then analysis of
only the clear supernatant.
&
\emph{Evidence-Decision Conflict.}
Excluded the solid fraction
required for representative
sampling.
\\
\bottomrule
\end{tabularx}
\caption{
Representative failures in model answers. The cases illustrate
misinterpretation of experimental evidence
and an operation inconsistent with the
sampling objective.
}
\label{tab:error-cases}
\end{table*}

\subsection{Analysis of Error Patterns}
\label{sec:error_patterns}

Finally, we analyzed non-full-score responses from the three top-performing models based on unrounded matched-image scores on the controlled organic-synthesis
subset: Claude Opus 4.7, GPT-5.5, and Gemini 3.1 Pro. The taxonomy in Appendix~\ref{app:error} defines six error types: Evidence Misreading, Evidence-Decision Conflict, Chemical Signal Errors, Common Rule Misuse, Mistaking Normal for Abnormal, and Experimental Method Errors.

Errors were concentrated in Evidence Misreading ($40.0$\%) and Evidence-Decision Conflict ($35.8$\%). Evidence Misreading occurred when models failed to identify, correctly read, or use decision-relevant experimental cues, whereas Evidence-Decision Conflict reflected explanations or recommendations inconsistent with the observed state or underlying cause. These two categories accounted for $75.8$\% of the error distribution, highlighting two related challenges: identifying the evidence that constrains an experimental judgment and producing conclusions consistent with the specific experimental situation. Common Rule Misuse accounted for a further $7.3$\%, reflecting the use of common laboratory rules in place of case-specific evidence and interpretation. This suggests that scientifically plausible responses may still lead to incorrect judgments when they overlook case-specific experimental evidence. These expert-identified patterns characterize errors in model outputs rather than directly establishing the internal mechanisms responsible for them.

Table~\ref{tab:error-cases} illustrates two types of
failures. In the TLC case, the model
misclassified standard spotting as non-standard and
attributed the pattern to sample overloading rather
than unsuitable solvent polarity, despite insufficient
supporting evidence. In the LC-MS case, the model
recommended centrifugation or filtration followed by
supernatant analysis, excluding the solid fraction
required for representative sampling before dissolution.
These cases show how plausible explanations or procedures
can lead to inappropriate judgments when inconsistent
with experimental evidence or objectives.
Further details are provided in Appendix~\ref{app:error}.

\section{Conclusion}
\label{sec:conclusion}

We introduced LabOPBench, a $756$-question benchmark for evidence-grounded laboratory decision-making, and evaluated eight current multimodal models. Overall normalized scores ranged from $28.8$\% to $45.8$\%. On the controlled $100$-question subset, the strongest models exceeded the undergraduate mean ($31.0$\%) but remained below the PhD mean ($45.9$\%). Image mismatch reduced the cross-model mean from $37.5$\% to $19.3$\%, indicating sensitivity to image-question alignment.

These results highlight a persistent gap between multimodal
perception and experimental decision-making. Models often
fail to correctly identify decision-relevant observations or reliably connect
them to experimental states, likely causes, and appropriate
actions. Evidence Misreading ($40.0\%$) and
Evidence-Decision Conflict ($35.8\%$) were the dominant
error types, together accounting for $75.8\%$ of the error distribution.

Future laboratory systems will require representations
linking observations to procedural context, recent
operations, and expected state transitions. They should also assess alternative interventions and determine when to act, take further measurements, or seek human review.
LabOPBench provides a basis for evaluating these
capabilities in increasingly adaptive laboratory workflows.
\section{Limitations}
\label{sec:limitations}

LabOPBench evaluates evidence-grounded laboratory judgment at a single decision point, given a fixed experimental context and multimodal evidence. It does not cover continuous experimental trajectories, post-action observations, robotic execution, or replanning after failed interventions. Accordingly, the findings should be interpreted as evidence of models' performance in static laboratory decision-making rather than as a complete evaluation of closed-loop experimental autonomy.
\bibliography{references}

\begin{thebibliography}{36}
\providecommand{\natexlab}[1]{#1}

\bibitem[{Alampara et~al.(2025)Alampara, Schilling-Wilhelmi,
  R\'{i}os-Garc\'{i}a, Mandal, Khetarpal, Grover, Krishnan, and
  Jablonka}]{alampara2025macbench}
Nawaf Alampara, Mara Schilling-Wilhelmi, Marti\~{n}o R\'{i}os-Garc\'{i}a,
  Indrajeet Mandal, Pranav Khetarpal, Hargun~Singh Grover, N.~M.~Anoop
  Krishnan, and Kevin~Maik Jablonka. 2025.
\newblock \href {https://doi.org/10.1038/s43588-025-00836-3} {Probing the
  limitations of multimodal language models for chemistry and materials
  research}.
\newblock \emph{Nature Computational Science}, 5:952--961.

\bibitem[{{Anthropic}(2026)}]{anthropic2026claudeopus47}
{Anthropic}. 2026.
\newblock \href {https://www.anthropic.com/claude-opus-4-7-system-card}
  {{Claude Opus 4.7 System Card}}.
\newblock System card, Anthropic.

\bibitem[{Boiko et~al.(2023)Boiko, MacKnight, Kline, and
  Gomes}]{boiko2023autonomous}
Daniil~A. Boiko, Robert MacKnight, Ben Kline, and Gabe Gomes. 2023.
\newblock \href {https://doi.org/10.1038/s41586-023-06792-0} {Autonomous
  chemical research with large language models}.
\newblock \emph{Nature}, 624(7992):570--578.

\bibitem[{Burger et~al.(2020)Burger, Maffettone, Gusev, Aitchison, Bai, Wang,
  Li, Alston, Li, Clowes, Rankin, Harris, Sprick, and
  Cooper}]{burger2020mobile}
Benjamin Burger, Phillip~M. Maffettone, Vladimir~V. Gusev, Catherine~M.
  Aitchison, Yang Bai, Xiaoyan Wang, Xiaobo Li, Ben~M. Alston, Buyi Li, Rob
  Clowes, Nicola Rankin, Brandon Harris, Reiner~Sebastian Sprick, and Andrew~I.
  Cooper. 2020.
\newblock \href {https://doi.org/10.1038/s41586-020-2442-2} {A mobile robotic
  chemist}.
\newblock \emph{Nature}, 583(7815):237--241.

\bibitem[{Coley et~al.(2019)Coley, Thomas, Lummiss, Jaworski, Breen, Schultz,
  Hart, Fishman, Rogers, Gao, Hicklin, Plehiers, Byington, Piotti, Green, Hart,
  Jamison, and Jensen}]{coley2019flow}
Connor~W. Coley, Dale~A. Thomas, Justin A.~M. Lummiss, Jonathan~N. Jaworski,
  Christopher~P. Breen, Victor Schultz, Travis Hart, Joshua~S. Fishman, Luke
  Rogers, Hanyu Gao, Robert~W. Hicklin, Pieter Plehiers, Joshua Byington,
  John~S. Piotti, William~H. Green, A.~John Hart, Timothy~F. Jamison, and
  Klavs~F. Jensen. 2019.
\newblock \href {https://doi.org/10.1126/science.aax1566} {A robotic platform
  for flow synthesis of organic compounds informed by {AI} planning}.
\newblock \emph{Science}, 365(6453):eaax1566.

\bibitem[{Cui et~al.(2025)Cui, Yao, Qin, Li, Wang, and
  Hu}]{cui2025chemolympiad}
Yiming Cui, Xin Yao, Yuxuan Qin, Xin Li, Shijin Wang, and Guoping Hu. 2025.
\newblock \href {https://doi.org/10.1038/s42004-025-01782-x} {Evaluating large
  language models on multimodal chemistry olympiad exams}.
\newblock \emph{Communications Chemistry}, 8:402.

\bibitem[{Darvish et~al.(2025)Darvish, Skreta, Zhao, Yoshikawa, Som,
  Bogdanovic, Cao, Hao, Xu, Aspuru-Guzik, Garg, and
  Shkurti}]{darvish2025organa}
Kourosh Darvish, Marta Skreta, Yuchi Zhao, Naruki Yoshikawa, Sagnik Som,
  Miroslav Bogdanovic, Yang Cao, Han Hao, Haoping Xu, Al{\'a}n Aspuru-Guzik,
  Animesh Garg, and Florian Shkurti. 2025.
\newblock \href {https://doi.org/10.1016/j.matt.2024.10.015} {{ORGANA}: A
  robotic assistant for automated chemistry experimentation and
  characterization}.
\newblock \emph{Matter}, 8(2):101897.

\bibitem[{El-Khawaldeh et~al.(2024{\natexlab{a}})El-Khawaldeh, Guy, Bork,
  Taherimakhsousi, Jones, Hawkins, Han, Pritchard, Cole, Monfette, and
  Hein}]{elkhawaldeh2024eye}
Rama El-Khawaldeh, Mason Guy, Finn Bork, Nina Taherimakhsousi, Kris~N. Jones,
  Joel~M. Hawkins, Lu~Han, Robert~P. Pritchard, Blaine~A. Cole, Sebastien
  Monfette, and Jason~E. Hein. 2024{\natexlab{a}}.
\newblock \href {https://doi.org/10.1039/D3SC05491H} {Keeping an ``eye'' on the
  experiment: Computer vision for real-time monitoring and control}.
\newblock \emph{Chemical Science}, 15(4):1271--1282.

\bibitem[{El-Khawaldeh et~al.(2024{\natexlab{b}})El-Khawaldeh, Mandal,
  Yoshikawa, Zhang, Corkery, Prieto, Aspuru-Guzik, Darvish, and
  Hein}]{elkhawaldeh2024lle}
Rama El-Khawaldeh, Abhijoy Mandal, Naruki Yoshikawa, Wenyu Zhang, Ryan Corkery,
  Paloma Prieto, Al{\'a}n Aspuru-Guzik, Kourosh Darvish, and Jason~E. Hein.
  2024{\natexlab{b}}.
\newblock \href {https://doi.org/10.1016/j.device.2024.100404} {From eyes to
  cameras: Computer vision for high-throughput liquid--liquid separation}.
\newblock \emph{Device}, 2(7):100404.

\bibitem[{Eppel et~al.(2020)Eppel, Xu, Bismuth, and
  Aspuru-Guzik}]{eppel2020labpics}
Sagi Eppel, Haoping Xu, Mor Bismuth, and Alan Aspuru-Guzik. 2020.
\newblock \href {https://doi.org/10.1021/acscentsci.0c00460} {Computer vision
  for recognition of materials and vessels in chemistry lab settings and the
  {Vector-LabPics} data set}.
\newblock \emph{ACS Central Science}, 6(10):1743--1752.

\bibitem[{{Google DeepMind}(2026{\natexlab{a}})}]{google2026gemini31flashlite}
{Google DeepMind}. 2026{\natexlab{a}}.
\newblock \href
  {https://deepmind.google/models/model-cards/gemini-3-1-flash-lite/} {{Gemini
  3.1 Flash-Lite Model Card}}.
\newblock Technical report, Google DeepMind.

\bibitem[{{Google DeepMind}(2026{\natexlab{b}})}]{google2026gemini31pro}
{Google DeepMind}. 2026{\natexlab{b}}.
\newblock \href {https://deepmind.google/models/model-cards/gemini-3-1-pro/}
  {{Gemini 3.1 Pro Model Card}}.
\newblock Technical report, Google DeepMind.

\bibitem[{Guo et~al.(2023)Guo, Guo, Nan, Liang, Guo, Chawla, Wiest, and
  Zhang}]{guo2023chemllmbench}
Taicheng Guo, Kehan Guo, Bozhao Nan, Zhenwen Liang, Zhichun Guo, Nitesh~V.
  Chawla, Olaf Wiest, and Xiangliang Zhang. 2023.
\newblock \href
  {https://proceedings.neurips.cc/paper_files/paper/2023/hash/bbb330189ce02be00cf7346167028ab1-Abstract-Datasets_and_Benchmarks.html}
  {What can large language models do in chemistry? a comprehensive benchmark on
  eight tasks}.
\newblock In \emph{Advances in Neural Information Processing Systems},
  volume~36, pages 59662--59688.

\bibitem[{H{\"a}se et~al.(2019)H{\"a}se, Roch, and Aspuru-Guzik}]{hase2019sdl}
Florian H{\"a}se, Lo{\"i}c~M. Roch, and Al{\'a}n Aspuru-Guzik. 2019.
\newblock \href {https://doi.org/10.1016/j.trechm.2019.02.007} {Next-generation
  experimentation with self-driving laboratories}.
\newblock \emph{Trends in Chemistry}, 1(3):282--291.

\bibitem[{Huang et~al.(2026)Huang, Zhang, He, Zhi, Wang, Chen, Liu, Li, Xu,
  Liu, Liang, Li, Cui, Xu, Wang, Liu, Lian, Liu, and Chen}]{huang2026chemeval}
Yuqing Huang, Rongyang Zhang, Xuesong He, Xuyang Zhi, Hao Wang, Nuo Chen,
  Zongbo Liu, Xin Li, Feiyang Xu, Deguang Liu, Huadong Liang, Yi~Li, Jian Cui,
  Yin Xu, Shijin Wang, Qi~Liu, Defu Lian, Guiquan Liu, and Enhong Chen. 2026.
\newblock \href {https://openreview.net/forum?id=JrqjSkEPrX} {{ChemEval}: A
  multi-level and fine-grained chemical capability evaluation for large
  language models}.
\newblock In \emph{The Fourteenth International Conference on Learning
  Representations}.

\bibitem[{Lai et~al.(2025)Lai, Zheng, Cai, Lyu, Yang, Liang, Hu, and
  Wang}]{lai2025matcha}
Zhengzhao Lai, Youbin Zheng, Zhenyang Cai, Haonan Lyu, Jingpu Yang, Hong-Qing
  Liang, Yan Hu, and Benyou Wang. 2025.
\newblock \href {https://doi.org/10.18653/v1/2025.findings-emnlp.235} {Can
  multimodal {LLM}s see materials clearly? a multimodal benchmark on materials
  characterization}.
\newblock In \emph{Findings of the Association for Computational Linguistics:
  EMNLP 2025}, pages 4384--4404, Suzhou, China. Association for Computational
  Linguistics.

\bibitem[{Laurent et~al.(2024)Laurent, Janizek, Ruzo, Hinks, Hammerling,
  Narayanan, Ponnapati, White, and Rodriques}]{laurent2024labbench}
Jon~M. Laurent, Joseph~D. Janizek, Michael Ruzo, Michaela~M. Hinks, Michael~J.
  Hammerling, Siddharth Narayanan, Manvitha Ponnapati, Andrew~D. White, and
  Samuel~G. Rodriques. 2024.
\newblock \href {https://arxiv.org/abs/2407.10362} {{LAB-Bench}: Measuring
  capabilities of language models for biology research}.
\newblock \emph{Preprint}, arXiv:2407.10362.

\bibitem[{Li et~al.(2025{\natexlab{a}})Li, Fang, Li, Huang, Wang, Wang, Bai,
  Hao, Lin, Liang, Zhang, and Ke}]{li2025rxnbench}
Hanzheng Li, Xi~Fang, Yixuan Li, Chaozheng Huang, Junjie Wang, Xi~Wang, Hongzhe
  Bai, Bojun Hao, Shenyu Lin, Huiqi Liang, Linfeng Zhang, and Guolin Ke.
  2025{\natexlab{a}}.
\newblock \href {https://doi.org/10.48550/arXiv.2512.23565} {{RxnBench}: A
  multimodal benchmark for evaluating large language models on chemical
  reaction understanding from scientific literature}.
\newblock \emph{arXiv preprint arXiv:2512.23565}.

\bibitem[{Li et~al.(2025{\natexlab{b}})Li, Zhang, Wang, Hao, Lei, Tan, Zhou,
  Liu, Yang, Xiong, Wang, Chen, Wang, Li, Su, Zhang, Ouyang, Li, and
  Zhou}]{li2025chemvlm}
Junxian Li, Di~Zhang, Xunzhi Wang, Zeying Hao, Jingdi Lei, Qian Tan, Cai Zhou,
  Wei Liu, Yaotian Yang, Xinrui Xiong, Weiyun Wang, Zhe Chen, Wenhai Wang, Wei
  Li, Mao Su, Shufei Zhang, Wanli Ouyang, Yuqiang Li, and Dongzhan Zhou.
  2025{\natexlab{b}}.
\newblock \href {https://doi.org/10.1609/aaai.v39i1.32020} {{ChemVLM}:
  Exploring the power of multimodal large language models in chemistry area}.
\newblock In \emph{Proceedings of the AAAI Conference on Artificial
  Intelligence}, volume~39, pages 415--423.

\bibitem[{Lin et~al.(2025{\natexlab{a}})Lin, Chen, Jia, Wang, Wang, Zhang,
  Ding, Du, Ding, and Liu}]{lin2025sdldataset}
Shiwei Lin, Zesen Chen, Xiaobin Jia, Yi~Wang, Chenxu Wang, Shuyuan Zhang,
  Xiaozhen Ding, Boyuan Du, Weili Ding, and Huaping Liu. 2025{\natexlab{a}}.
\newblock \href {https://doi.org/10.1038/s41597-025-06060-y} {A visual dataset
  for anomaly detection in self-driving laboratories}.
\newblock \emph{Scientific Data}, 12(1):1787.

\bibitem[{Lin et~al.(2025{\natexlab{b}})Lin, Wang, Ding, Wang, Du, Song, Wang,
  and Liu}]{lin2025vlmanomaly}
Shiwei Lin, Chenxu Wang, Xiaozhen Ding, Yi~Wang, Boyuan Du, Lei Song, Chenggang
  Wang, and Huaping Liu. 2025{\natexlab{b}}.
\newblock \href {https://doi.org/10.1109/ICARM65671.2025.11293532} {A
  {VLM}-based method for visual anomaly detection in robotic scientific
  laboratories}.
\newblock In \emph{2025 10th IEEE International Conference on Advanced Robotics
  and Mechatronics ({ICARM})}, pages 34--39. IEEE.

\bibitem[{Lu et~al.(2024)Lu, Cao, Liu, Bai, Chen, Yao, Zheng, and
  Li}]{lu2024moleculeqa}
Xingyu Lu, He~Cao, Zijing Liu, Shengyuan Bai, Leqing Chen, Yuan Yao, Hai-Tao
  Zheng, and Yu~Li. 2024.
\newblock \href {https://doi.org/10.18653/v1/2024.findings-emnlp.216}
  {{MoleculeQA}: A dataset to evaluate factual accuracy in molecular
  comprehension}.
\newblock In \emph{Findings of the Association for Computational Linguistics:
  EMNLP 2024}, pages 3769--3789, Miami, Florida, USA. Association for
  Computational Linguistics.

\bibitem[{{MiniMax}(2026)}]{minimax2026m3}
{MiniMax}. 2026.
\newblock \href {https://www.minimax.io/blog/minimax-m3} {{MiniMax M3: Frontier
  Coding, 1M Context, Native Multimodality --- All in One Model}}.
\newblock MiniMax Research.

\bibitem[{Mirza et~al.(2025)Mirza, Alampara, Kunchapu, R\'{i}os-Garc\'{i}a,
  Emoekabu, Krishnan, Gupta, Schilling-Wilhelmi, Okereke, Aneesh, Asgari,
  Eberhardt, Elahi, Elbeheiry, Gil, Glaubitz, Greiner, Holick, Hoffmann,
  Ibrahim, Klepsch, K\"{o}ster, Kreth, Meyer, Miret, Peschel, Ringleb, Roesner,
  Schreiber, Schubert, Stafast, Dinga~Wonanke, Pieler, Schwaller, and
  Jablonka}]{mirza2025chembench}
Adrian Mirza, Nawaf Alampara, Sreekanth Kunchapu, Marti\~{n}o
  R\'{i}os-Garc\'{i}a, Benedict Emoekabu, Aswanth Krishnan, Tanya Gupta, Mara
  Schilling-Wilhelmi, Macjonathan Okereke, Anagha Aneesh, Mehrdad Asgari,
  Juliane Eberhardt, Amir~Mohammad Elahi, Hani~M. Elbeheiry, Mar\'{i}a~Victoria
  Gil, Christina Glaubitz, Maximilian Greiner, Caroline~T. Holick, Tim
  Hoffmann, and 16 others. 2025.
\newblock \href {https://doi.org/10.1038/s41557-025-01815-x} {A framework for
  evaluating the chemical knowledge and reasoning abilities of large language
  models against the expertise of chemists}.
\newblock \emph{Nature Chemistry}, 17(7):1027--1034.

\bibitem[{{OpenAI}(2026)}]{openai2026gpt55}
{OpenAI}. 2026.
\newblock \href {https://openai.com/index/gpt-5-5-system-card/} {{GPT-5.5
  System Card}}.
\newblock System card, OpenAI.

\bibitem[{{Qwen Team}(2026)}]{qwenteam2026qwen36plus}
{Qwen Team}. 2026.
\newblock \href {https://qwen.ai/blog?id=qwen3.6} {{Qwen3.6-Plus: Towards Real
  World Agents}}.
\newblock Qwen technical blog.

\bibitem[{Runcie et~al.(2026)Runcie, Deane, and Imrie}]{runcie2026chemiq}
Nicholas~T. Runcie, Charlotte~M. Deane, and Fergus Imrie. 2026.
\newblock \href {https://doi.org/10.1021/acs.jcim.5c02145} {Assessing the
  chemical intelligence of large language models}.
\newblock \emph{Journal of Chemical Information and Modeling}, 66(1):216--227.

\bibitem[{Tom et~al.(2024)Tom, Schmid, Baird, Cao, Darvish, Hao, Lo,
  Pablo-Garc{\'i}a, Rajaonson, Skreta, Yoshikawa, Corapi, Akkoc,
  Strieth-Kalthoff, Seifrid, and Aspuru-Guzik}]{tom2024sdl}
Gary Tom, Stefan~P. Schmid, Sterling~G. Baird, Yang Cao, Kourosh Darvish, Han
  Hao, Stanley Lo, Sergio Pablo-Garc{\'i}a, Ella~M. Rajaonson, Marta Skreta,
  Naruki Yoshikawa, Samantha Corapi, Gun~Deniz Akkoc, Felix Strieth-Kalthoff,
  Martin Seifrid, and Al{\'a}n Aspuru-Guzik. 2024.
\newblock \href {https://doi.org/10.1021/acs.chemrev.4c00055} {Self-driving
  laboratories for chemistry and materials science}.
\newblock \emph{Chemical Reviews}, 124(16):9633--9732.

\bibitem[{Tong et~al.(2024)Tong, Liu, Zhai, Ma, LeCun, and
  Xie}]{tong2024eyeswide}
Shengbang Tong, Zhuang Liu, Yuexiang Zhai, Yi~Ma, Yann LeCun, and Saining Xie.
  2024.
\newblock Eyes wide shut? exploring the visual shortcomings of multimodal llms.
\newblock In \emph{Proceedings of the IEEE/CVF Conference on Computer Vision
  and Pattern Recognition (CVPR)}, pages 9568--9578.

\bibitem[{{xAI}(2026)}]{xai2026grok43}
{xAI}. 2026.
\newblock \href {https://docs.x.ai/developers/models/grok-4.3} {{Grok 4.3}}.
\newblock xAI developer documentation.

\bibitem[{Xie et~al.(2025)Xie, Wang, Gao, Yang, Wan, Zhang, Fu, and
  Li}]{xie2025qcbench}
Jiaqing Xie, Weida Wang, Ben Gao, Zhuo Yang, Haiyuan Wan, Shufei Zhang, Tianfan
  Fu, and Yuqiang Li. 2025.
\newblock \href {https://doi.org/10.1021/acs.jcim.5c02033} {{QCBench}:
  Evaluating large language models on domain-specific quantitative chemistry}.
\newblock \emph{Journal of Chemical Information and Modeling},
  65(22):12268--12278.

\bibitem[{Yue et~al.(2025)Yue, Zheng, Ni, Wang, Zhang, Tong, Sun, Yu, Zhang,
  Sun, Su, Chen, and Neubig}]{yue2025mmmupro}
Xiang Yue, Tianyu Zheng, Yuansheng Ni, Yubo Wang, Kai Zhang, Shengbang Tong,
  Yuxuan Sun, Botao Yu, Ge~Zhang, Huan Sun, Yu~Su, Wenhu Chen, and Graham
  Neubig. 2025.
\newblock \href {https://doi.org/10.18653/v1/2025.acl-long.736} {{MMMU}-pro: A
  more robust multi-discipline multimodal understanding benchmark}.
\newblock In \emph{Proceedings of the 63rd Annual Meeting of the Association
  for Computational Linguistics (Volume 1: Long Papers)}, pages 15134--15186,
  Vienna, Austria. Association for Computational Linguistics.

\bibitem[{Zhang et~al.(2026)Zhang, Liang, Zhang, Yu, Yang, Jin, and
  Xu}]{zhang-etal-2026-chemreason}
Jinwei Zhang, Xucheng Liang, Yu~Zhang, Ruijie Yu, Xiaokang Yang, Yaohui Jin,
  and Yanyan Xu. 2026.
\newblock \href {https://doi.org/10.18653/v1/2026.acl-long.1535}
  {{C}hem{R}eason-bench: Benchmarking large language models for procedural
  reasoning in experimental chemistry}.
\newblock In \emph{Proceedings of the 64th Annual Meeting of the Association
  for Computational Linguistics (Volume 1: Long Papers)}, pages 33211--33248.
  Association for Computational Linguistics.

\bibitem[{Zhou et~al.(2025)Zhou, Cheng, Mao, Luo, Liu, Li, Zhang, Liu, Li, and
  Chen}]{zhou2025chemtable}
Yitong Zhou, Mingyue Cheng, Qingyang Mao, Yucong Luo, Qi~Liu, Yupeng Li,
  Xiaohan Zhang, Deguang Liu, Xin Li, and Enhong Chen. 2025.
\newblock \href {https://doi.org/10.48550/arXiv.2506.11375} {Benchmarking
  multimodal llms on recognition and understanding over chemical tables}.
\newblock \emph{arXiv preprint arXiv:2506.11375}.

\bibitem[{Zhou et~al.(2026)Zhou, Yang, Huang, Guo, Emory, Ghosh, Bedar, Shekar,
  Liang, Chen, Gao, Geyer, Moniz, Chawla, and Zhang}]{zhou2026labsafety}
Yujun Zhou, Jingdong Yang, Yue Huang, Kehan Guo, Zoe Emory, Bikram Ghosh, Amita
  Bedar, Sujay Shekar, Zhenwen Liang, Pin-Yu Chen, Tian Gao, Werner Geyer, Nuno
  Moniz, Nitesh~V. Chawla, and Xiangliang Zhang. 2026.
\newblock \href {https://doi.org/10.1038/s42256-025-01152-1} {Benchmarking
  large language models on safety risks in scientific laboratories}.
\newblock \emph{Nature Machine Intelligence}, 8:20--31.

\bibitem[{Zou et~al.(2026)Zou, Zhu, Zhu et~al.}]{zou2026interns1pro}
Yicheng Zou, Dongsheng Zhu, Lin Zhu, et~al. 2026.
\newblock \href {https://arxiv.org/abs/2603.25040} {{Intern-S1-Pro: Scientific
  Multimodal Foundation Model at Trillion Scale}}.
\newblock \emph{arXiv preprint arXiv:2603.25040}.

\end{thebibliography}
\definecolor{jsonbg}{HTML}{FAFBFC}
\definecolor{jsonborder}{HTML}{DDE2E7}
\definecolor{jsonstring}{HTML}{356783}
\definecolor{jsonkeyword}{HTML}{965A38}

\lstdefinelanguage{SIJSON}{
    sensitive=true,
    morestring=[b]",
    morekeywords={true,false,null},
    stringstyle=\color{jsonstring},
    keywordstyle=\color{jsonkeyword}
}

\lstdefinestyle{SIjson}{
    language=SIJSON,
    basicstyle=\ttfamily\footnotesize,
    backgroundcolor=\color{jsonbg},
    frame=single,
    rulecolor=\color{jsonborder},
    framerule=0.4pt,
    framesep=6pt,
    numbers=none,
    showstringspaces=false,
    breaklines=true,
    breakatwhitespace=true,
    breakindent=1em,
    columns=fullflexible,
    keepspaces=true,
    tabsize=2,
    aboveskip=8pt,
    belowskip=8pt
}


\clearpage
\appendix
\setcounter{secnumdepth}{3}
\section{Appendix}
\label{sec:appendix}

\subsection{Standardized Question-Authoring Protocol}
\label{app:question-construction}
This protocol is intended to guide experts in constructing initial benchmark questions based on authentic chemical laboratory materials. Each question should assess the model's ability to integrate visual evidence, instrumental information, or experimental observations within a specific experimental context to perform state identification, cause analysis, result interpretation, risk assessment, or operational decision-making.

Questions must not merely assess textbook knowledge detached from experimental scenarios, nor should they depend on information that is not provided in the image, question prompt, or supplied background. Each question must have a clearly defined experimental judgment target and must be supportable by the provided evidence together with relevant domain knowledge. This protocol covers only the construction of the initial question draft. After completion, each question will proceed to independent expert review, revision, and, when necessary, third-party adjudication.
Each question is assigned only one primary category to ensure mutual exclusivity of categories in statistical analysis and evaluation. Question materials should satisfy the following requirements:

\begin{itemize}
\item Images, instrument outputs, or experimental observations must be directly relevant to the judgment target being assessed.
\item Key visual or experimental cues should be reasonably observable and should not depend on difficult-to-discern minor details, unavailable preceding or subsequent context, or the author's personal memory.
\item Image cropping, rotation, de-identification, or resolution adjustment must not alter the original experimental facts or obscure key evidence.
\item Images may show normal operations, abnormal states, incorrect settings, instrument outputs, experimental results, or representative process phenomena, but a question must not be constructed merely because an image ``looks abnormal.''
\end{itemize}

\subsubsection{Organic Synthesis}
Organic-synthesis questions are organized according to authentic experimental workflows into four modules and sixteen subcategories (Table~\ref{tab:si-1}).

\begin{table}[!htbp]

\centering
{\small
\setlength{\tabcolsep}{4.5pt}
\renewcommand{\arraystretch}{1.16}
\begin{tabularx}{\linewidth}{@{}>{\raggedright\arraybackslash\hsize=0.7200\hsize}X>{\raggedright\arraybackslash\hsize=1.2800\hsize}X@{}}
\toprule
\textbf{categories} & \textbf{Subcategories} \\
\midrule
Experiment Preparation & Experiment\_plan; Experiment\_Safety \\
Experiment Monitor & Experiment\_Monitor; Experiment\_notes \\
Experiment Postprocess & Extraction; Evaporation; Column; Recrystallization; Trituration; TLC; Lyophilization \\
Experiment Characterization & NMR; LCMS; Meltingpoint; Crystal; GCMS \\
\bottomrule
\end{tabularx}
}

\caption{Organic-synthesis categories and subcategories.}\label{tab:si-1}
\end{table}

\subsubsection{Materials Science}

Materials science questions are uniformly labeled as Materials Science. The question author should briefly describe the experimental scenario in the question metadata, such as materials preparation, testing conditions, microscopic observation, structural or performance characterization, interpretation of instrument outputs, or analysis of experimental abnormalities.
Materials-science questions should be organized according to their own experimental scenarios and are not required to use the four-stage labels defined for organic synthesis.

\subsection{ Question-Design Principles}
Questions should be centered on key evidence contained in the provided image or experimental information. The question author should be able to clearly answer:
1. What key evidence must the model identify?
2. What experimental judgment does this evidence support?
3. Does this judgment require integration with the provided experimental background?
4. Are there other equally reasonable interpretations?

When the available evidence is insufficient
for a specific causal attribution, the question
should focus on judgments supported by the
experimental context, such as identifying
the observed state or determining an
appropriate verification step.
Questions with unresolved ambiguity
should be revised or excluded.

\subsubsection{Single Core Judgment Target}
Multiple interrelated subquestions are permitted, but they must all serve the same experimental decision chain. For example, ``whether the observed phenomenon is sufficient to determine reaction completion---how it should be verified---what should be prioritized for optimization if an abnormality is present'' may be formulated as a continuous monitoring-decision question. Multiple independent knowledge questions must not be mechanically combined into a single question.
Each question should have one primary experimental judgment target, for example:
1.determining whether an operation is appropriate or poses a risk;
2.determining whether a reaction has reached the expected state;
3.identifying a possible cause of an abnormal observation;
4.selecting an appropriate workup, purification, or characterization operation;
5.interpreting an instrumental or characterization result;
6.proposing an actionable corrective measure that is consistent with the evidence.

\subsubsection{Authentic Experimental Context}
The question prompt should provide the minimum sufficient background required for answering, including the necessary experimental stage, object of operation, observed phenomenon, instrument type, or known conditions. Questions should not require the model to guess reagent identities, reaction types, concentrations, temperatures, instrument parameters, or previous operations that have not been provided.
Questions should primarily assess judgments made in authentic experimental settings rather than use generalized statements detached from a specific context. For example:

It is not appropriate to ask only, ``What causes tailing in TLC?''

Instead, the question should ask, ``Based on the provided TLC image, developing system, and sample state, how should the current spot abnormality be interpreted and how should the conditions be adjusted?''

\subsubsection{Evidence Sufficiency and Verification}
Color, turbidity, precipitation, foam, liquid-level changes, band morphology, or instrument readings may serve as important cues. However, unless such a phenomenon has sufficient and verifiable specificity within the given experimental system, it must not be used as the sole basis for determining product formation, reaction completion, or yield.
When additional verification is relevant to
the target judgment, appropriate analytical
methods, such as TLC, LCMS, GCMS, NMR, or
melting-point analysis, may be considered
within the experimental context.

\subsection{Required Content for Each Question}
Each expert author should submit the following content. The question should:
1.clearly describe the experimental scenario;
2.specify the judgment task that the model is required to perform;
3.provide the minimum necessary background required to make the judgment;
4.use accurate, actionable, and unambiguous experimental terminology;
5.avoid revealing the correct answer through wording, answer ordering, or obvious hints.

\noindent
\begin{minipage}{\linewidth}
\begin{lstlisting}[
    basicstyle=\ttfamily\footnotesize,
    breaklines=true,
    breakatwhitespace=true,
    columns=fullflexible,
    frame=single,
    framesep=5pt,
    xleftmargin=3pt,
    xrightmargin=3pt
]
{
  "category": "one primary category",
  "experimental_stage_or_context":
    "experimental stage or materials-science experimental context",
  "evidence_type":
    "operation image, instrument interface, spectrum, chromatogram, microscopy image, experimental observation, or other evidence type",
  "image":
    "corresponding image filename(s), which may contain a single image or multiple images",
  "target_decision":
    "the primary experimental judgment target assessed by the question"
}
\end{lstlisting}
\end{minipage}
\par

Each question should include an internally used description of key evidence, including:
1.the most important observable cues in the image or experimental record;
2.the relationship between those cues and the experimental judgment;
3.reasonable alternative explanations or alternative operations that should be accepted;
4.core incorrect judgments that should not be accepted.


\subsubsection{Reference Answer and Tiered Scoring Anchors}

\begin{table*}[t]

\centering
{\small
\setlength{\tabcolsep}{4.5pt}
\renewcommand{\arraystretch}{1.16}
\begin{tabularx}{\textwidth}{@{}>{\raggedright\arraybackslash\hsize=0.6800\hsize}X>{\centering\arraybackslash\hsize=0.3600\hsize}X>{\raggedright\arraybackslash\hsize=1.9600\hsize}X@{}}
\toprule
\textbf{Anchor} & \textbf{Score} & \textbf{Definition} \\
\midrule
Score 3 anchor & 3 & Complete, scientifically correct, consistent with the supplied evidence, and covers the core judgment, causal explanation, and necessary operational recommendations required by the question. \\
Score 2 anchor & 2 & The core judgment is correct, but key evidence, verification steps, causal explanation, or corrective measures are omitted; no substantive error that would change the conclusion should be present. \\
Score 1 anchor & 1 & Identifies some relevant observations or proposes a directionally reasonable recommendation, but does not complete the core judgment, lacks the necessary evidence chain, or uses an approach insufficient to support the conclusion. \\
Score 0 anchor & 0 & The core judgment is incorrect, contradicts the supplied evidence, proposes clearly inappropriate or unsafe operations, or is irrelevant to the question. \\
\bottomrule
\end{tabularx}
}

\caption{Four-level scoring anchors for benchmark questions.}\label{tab:si-2}
\end{table*}
The Score 3 / Score 2 / Score 1 / Score 0 anchors should reflect a monotonically decreasing scoring logic (Table~\ref{tab:si-2}). Lower-scoring answers should primarily differ through missing key information, incomplete evidence chains, or incorrect core conclusions.
In addition to the four scoring anchors, each question should include an internal scoring basis specifying at least:
1.the key judgment(s) that must be included in a full-score answer;
2.acceptable alternative phrasings, verification methods, or corrective measures;
3.key omissions that would result in substantial score deductions;
4.core errors or dangerous recommendations that would result in a score of zero.
If multiple scientifically equivalent answer pathways exist, they should be explicitly documented in the scoring basis to avoid treating a single wording or a single operational route as the only correct answer.

\subsection{Common Reasons for Rejection or Revision}
The Score 3 / Score 2 / Score 1 / Score 0 anchors should not contain evaluative statements such as ``this answer is incorrect,'' ``this answer lacks expertise,'' or ``this option may mislead the model.'' Such information should instead be placed in the internal scoring basis.
The following situations should be avoided during initial drafting or returned for revision during review:
1.the image is not directly related to the question prompt;
2.a deterministic conclusion is drawn solely from color, appearance, or a single observation beyond what the evidence can support;
3.the question depends on experimental conditions that have not been provided;
4.the visual evidence is insufficient to distinguish correct from incorrect answers;
5.multiple equally reasonable answers exist but are not covered by the scoring basis;
6.the question primarily tests textbook recall rather than using multimodal experimental information;
7.the Score 3 / Score 2 / Score 1 / Score 0 anchors do not exhibit a clear hierarchy of performance;
8.the full-score answer contains irrelevant, unverifiable, or overly specific operational recommendations;

\subsection{Expert Self-Check Checklist}
Before submitting an initial draft, confirm each of the following:

\begingroup
\setlength{\fboxsep}{7pt}
\noindent\fbox{%
\begin{minipage}{\dimexpr\linewidth-2\fboxsep-2\fboxrule\relax}
\small
\begin{itemize}
\setlength{\itemsep}{3pt}
\setlength{\topsep}{3pt}
\setlength{\parsep}{0pt}

\item Is the question based on an authentic experimental scenario or authentic experimental data?

\item Does the question have only one clearly defined primary experimental judgment target?

\item Can the model make the required judgment using only the provided image, question prompt, and necessary background?

\item Have the key visual or experimental cues been clearly identified?

\item Is the reference answer scientifically correct and consistent with the evidence?

\item Are there multiple reasonable answers? If so, have they been documented in the scoring basis?

\item Do the Score 3 / Score 2 / Score 1 / Score 0 anchors form a clear and monotonic scoring hierarchy?

\item Is the Score 2 anchor ``core judgment correct but incomplete,'' rather than containing a decisive error?

\item Does the Score 0 anchor contain a clear error in the core judgment or operational recommendation?

\item Has the question been accurately assigned to one primary category?

\item Does the question avoid revealing the correct answer through prompt wording, answer phrasing, or meta-commentary?

\item Can the question be independently reviewed by another expert with comparable operational experience?

\end{itemize}
\end{minipage}%
}
\endgroup
\par

\subsection{ Standard Submission Format}

\begin{lstlisting}[
    basicstyle=\ttfamily\footnotesize,
    breaklines=true,
    breakatwhitespace=true,
    columns=fullflexible,
    frame=single,
    framesep=5pt,
    xleftmargin=3pt,
    xrightmargin=3pt,
    aboveskip=6pt,
    belowskip=6pt
]
{
  "category": "Experiment_Monitor",
  "experimental_stage_or_context": "reaction monitoring",
  "evidence_type": [
    "reaction-image"
  ],
  "image": [
    "Experiment_Monitor_001_01.png"
  ],
  "target_decision":
    "determine whether visual appearance alone can establish reaction completion and identify appropriate verification and troubleshooting steps",
  "question": "...",
  "key_evidence": [
    "...",
    "..."
  ],
  "answers": [
    {
      "label": "Score 3",
      "score": 3,
      "answer": "..."
    },
    {
      "label": "Score 2",
      "score": 2,
      "answer": "..."
    },
    {
      "label": "Score 1",
      "score": 1,
      "answer": "..."
    },
    {
      "label": "Score 0",
      "score": 0,
      "answer": "..."
    }
  ]
}
\end{lstlisting}

\section{Contributor Qualifications and Backgrounds}
\label{app:contributors}
LabOPBench was constructed and reviewed by 22 contributors with hands-on laboratory experience relevant to the experimental tasks represented in the benchmark. The contributor pool included researchers and laboratory specialists at different career stages, providing experience across reaction setup, experimental monitoring, workup and purification, characterization, and laboratory troubleshooting.

\subsection{Contributor composition}
The 22 contributors comprised four PhD researchers, six postdoctoral researchers, eight laboratory specialists, and four associate researchers or above. The information is summarized in Table~\ref{tab:si-3}.


\begin{table}[!htbp]
\centering
\begingroup
\small
\setlength{\tabcolsep}{9pt}
\renewcommand{\arraystretch}{1.08}
\setlength{\arrayrulewidth}{0.5pt}

\begin{tabular}{@{}l|c@{}}
\hline
\textbf{Position} & \textbf{Contributors} \\
\hline
PhD researcher & 4 \\
Postdoctoral researcher & 6 \\
Laboratory specialist & 8 \\
Associate researcher or above & 4 \\
\hline
\textbf{Total} & \textbf{22} \\
\hline
\end{tabular}

\endgroup
\caption{Contributor composition by position.}
\label{tab:si-3}
\end{table}

All contributors had at least 3 years of laboratory experience. Two contributors had 3--5 years of experience, eight had 5--7 years, four had 7--10 years, and eight had more than 10 years of laboratory experience. Overall, 20 of the 22 contributors had at least 5 years of laboratory experience, 12 had at least 7 years, and 8 had more than 10 years of experience. The information is summarized in Table~\ref{tab:si-4}.


\begin{table}[!htbp]
\centering
\begingroup
\small
\setlength{\tabcolsep}{10pt}
\renewcommand{\arraystretch}{1.08}
\setlength{\arrayrulewidth}{0.5pt}

\begin{tabular}{@{}l|c|c@{}}
\hline
\textbf{Laboratory experience} &
\textbf{Contributors} &
\textbf{Proportion} \\
\hline
3--5 years  & 2 & 9.1\% \\
5--7 years  & 8 & 36.4\% \\
7--10 years & 4 & 18.2\% \\
$>10$ years & 8 & 36.4\% \\
\hline
\textbf{Total} & \textbf{22} & \textbf{100.0\%} \\
\hline
\end{tabular}

\endgroup
\caption{Contributor laboratory experience.}
\label{tab:si-4}
\end{table}

Contributors were selected on the basis of hands-on laboratory experience relevant to the experimental situations represented in LabOPBench. Collectively, their expertise covered reaction setup, experimental monitoring, workup and purification, characterization, instrument operation, and troubleshooting. This range of experience supported both the construction of laboratory-grounded questions and the independent assessment of their scientific correctness, evidence sufficiency, and experimental relevance.


\section{Model Evaluation Setup and Parameters}
\label{app:Evaluation}
Eight multimodal models were evaluated using an OpenAI-compatible Chat Completions interface. All benchmark questions were evaluated independently using a common input format and answer-generation instruction. The evaluation framework supplied the question and associated visual evidence to the answer models without reference answers or scoring rubrics.
Each benchmark question was evaluated in an independent API request without conversational history. No system message was used. For multimodal questions, a single user message contained the following components in order: one image\_url block for each associated image;
 the question text; and the common answer-generation instruction.

Image filenames and local image paths appearing in the question text were replaced with neutral references such as ``Figure 1'' before submission. The question identifier, category, reference answers and question-specific scoring rubric were not provided to the answer model.
Images were read from the local image directory, Base64 encoded and transmitted as data URLs in the following format:

\noindent\texttt{\detokenize{data:{MIME type};base64,{encoded image}}}

\subsection{Answer prompt}
All eight answer models received the same instruction, presented below in a structured format.

\begin{lstlisting}[
    basicstyle=\ttfamily\footnotesize,
    breaklines=true,
    breakatwhitespace=true,
    columns=fullflexible,
    frame=single,
    framesep=5pt,
    xleftmargin=3pt,
    xrightmargin=3pt,
    aboveskip=6pt,
    belowskip=6pt
]
{
  "instruction": "Please carefully analyze the question and images (if any) above, and provide your answer.",
  "requirements": [
    "1. If the question contains multiple sub-questions, address EACH one individually.",
    "2. For each sub-question, identify and provide MAIN cause and its corresponding optimal solution. DO NOT list multiple causes, multiple solutions, or redundant details for any single sub-question.",
    "3. Your answer should be accurate and professional.",
    "4. Output the answer directly and concisely.",
    "5. If the question asks \"why\", provide only the core reason together with the answer, no extra content."
  ]
}
\end{lstlisting}

Answer generation was attempted up to ten times for each question, with a fixed 3-s delay after an unsuccessful attempt. No exponential backoff, jitter or explicit request timeout was configured.
An empty model response was treated as a failed attempt and retried. If all ten attempts failed, the corresponding output record contained predicted\_output=null and error=true. Failure of an individual question did not terminate evaluation of the remaining benchmark questions, and records for which answer generation ultimately failed were not subsequently scored.

\section{Automated Scoring Method and Prompts}
\label{app:score}
Model responses were independently evaluated by two judge models, GPT-5.5 (gpt-5.5) and Gemini 3.1 Pro (gemini-3.1-pro-preview). Each judge evaluated the candidate response against the question-specific 3-, 2-, 1-, and 0-point rubric levels. The two judge-specific evaluation results were retained separately and subsequently combined to obtain the reported score.
\subsection{Scoring framework}
The scoring framework used a 0--3 scale. The default procedure was holistic rubric matching, in which the candidate answer was matched to the single rubric level that it best satisfied.
Split scoring was used only when the original question explicitly contained multiple independent subquestions. For explicitly numbered subquestions, split scoring was mandatory. Each subquestion was scored independently, and the resulting subquestion scores were averaged and rounded to the nearest 0.5.
For each holistic answer or explicit subquestion, the judge was instructed to evaluate the first substantive claim or first complete procedure that directly answered the question. Introductory material, later alternative explanations, alternative solutions, and redundant details were ignored unless they contradicted the initially scored claim, changed the final recommendation, or introduced a dangerous operation.
Scores of 3, 2, 1, and 0 were assigned by matching the response to the scenario-specific rubric rather than by isolated keyword overlap. Half-point scores were permitted under the conditions defined in the scoring prompt: 2.5 when the response fully satisfied the 2-point rubric level and partially satisfied a key differentiating requirement of the 3-point level; 1.5 when the response fully satisfied the 1-point rubric level and partially satisfied a key differentiating requirement of the 2-point level; 0.5 when the response was relevant and experimentally reasonable but did not clearly satisfy any positive rubric level.

\subsection{Complete automated-scoring prompt}
\label{app:scoring-prompt}

Both judge models received the following scoring prompt:

\begin{lstlisting}[style=SIjson]
{
" instructions": [
  "You are a senior laboratory evaluator in experimental sciences.",
  "Your task is to score a model-generated experimental response against the provided A-D rubric options.",
  "The rubric options are authoritative. Do not replace them with your own general laboratory judgment."
],

"## Original question:": "{question_section}",
"## Student answer:": "{answer}",
"## Ground truth and scoring rubric options:": "{options_text}",

"Scoring rules": {

"Step 1 - Decide the scoring mode": [
  "Default mode is holistic A-D matching: choose the single rubric option whose complete answer is best matched by the student answer.",
  "Use split scoring only when the original question explicitly asks multiple independent sub-questions, such as numbered questions \"1, 2, 3\" or clearly separate tasks.",
  "If the original question explicitly contains numbered sub-questions, split scoring is mandatory. Do not use holistic scoring merely because the rubric options are written as integrated answers.",
  "Do not create scoring units from details inside the rubric. Scoring units may only come from the original question.",
  "If the question asks for a cause plus a solution, or a diagnosis plus an operation, treat them as one holistic answer unless the question explicitly separates them into independent sub-questions."
],

"Step 2 - Apply the first-point rule": [
  "For each holistic answer or explicit sub-question, score only the first substantive claim or first complete procedure that directly answers it.",
  "Ignore introductions, background comments, later alternative causes, later alternative solutions, and redundant details.",
  "Later information may affect the score only if it contradicts the first scored claim, changes the final recommendation, or introduces a dangerous operation."
],

"Step 3 - Match to the A-D rubric": [
  "Match the first scored content to the rubric option A/B/C/D by scenario-specific meaning, not by isolated keywords.",
  "Treat each rubric option as a complete condition-action package, not as a menu of independent keywords. The score should reflect the best complete package matched by the answer.",
  "A score of 3 requires matching the key differentiating details of the 3-point option.",
  "A higher score requires matching all key differentiating components of that rubric option when the option explicitly contains multiple required components, such as observation or diagnosis, cause, and required operation.",
  "If the answer matches only one component, such as only the cause or only one operation parameter, do not assign the full score for that rubric option.",
  "A score of 2 requires matching the key differentiating details of the 2-point option, without satisfying the 3-point differentiators.",
  "A score of 1 requires matching the 1-point option or a clearly valid low-level scoring element supported by the rubric.",
  "A score of 0 is required if the answer is irrelevant, opposite to the correct conclusion, unsafe, matches the 0-point option, or provides no score-bearing information.",
  "If an answer is generally scientifically plausible but misses the scenario-specific key point required by the rubric, it cannot receive 2 or 3 points. It may receive at most 0.5 limited credit.",
  "Do not upgrade a score because the answer sounds reasonable in general. The rubric is more important than general plausibility.",
  "Do not upgrade a score merely because the answer shares a broad goal, general mechanism, or laboratory category with a higher-score option. It must match the higher-score option's specific diagnostic premise and required operation.",
  "If the rubric specifies a concrete required operation, a generic operation category or a different operation with a similar intended effect is not enough for full credit unless the rubric explicitly allows such substitutions.",

  "Visual Evidence Rule:",
  "  - Do not require the answer to restate visual details merely because an image is present.",
  "  - Treat a visual observation as score-bearing only when the question explicitly asks for it, or when the rubric uses it to distinguish between score levels.",
  "  - If the answer matches the required diagnosis and operation but omits a non-contradictory visual observation required by the target higher-score rubric level, cap the score at the next lower full-score level.",
  "    For example, a 3-point option is capped at 2.0, and a 2-point option is capped at 1.0.",
  "  - If the answer contradicts a visual observation that is central to the rubric distinction, do not assign the contradicted higher-score level. Score it according to the lower rubric level it best matches.",
  "  - If the visual observation is only background context and not a differentiating rubric component, do not penalize the omission.",

  "If the rubric's higher-score option depends on a specific physical state or material form, the answer must match that state for full credit. If the answer replaces it with a different physical state or material form, do not assign that higher-score level; score it according to the lower rubric level it best matches, with a maximum of 1.0 if no lower level clearly matches.",
  "Do not treat different scenario-critical diagnoses as equivalent unless the rubric explicitly says they are equivalent.",
  "Scenario-critical diagnoses include, but are not limited to: physical phenomena such as oiling out, precipitation, emulsion, crystallization, phase separation, undissolved solid, and dark mother liquor; chromatography and TLC issues such as tailing, overloading, adsorption, solvent-polarity mismatch, decomposition, poor separation, fronting, saturation, and baseline drift; spectroscopy and MS assignments such as solvent peak, impurity peak, product ion, fragment ion, adduct, isotope peak, co-elution, wrong molecular weight, and wrong structural assignment; safety and operation diagnoses such as air leakage, water ingress, overheating, wrong apparatus setup, wrong layer selection, and unsafe reagent handling.",
  "If the first diagnosis identifies a different scenario-critical problem from the target rubric, the score for that holistic answer or sub-question cannot exceed 1.0, even if later cause or solution details overlap with a higher-score option."
],

"Step 4 - Half-point rule": [
  "Use half-point scores sparingly.",
  "Assign 2.5 only when the answer fully satisfies the 2-point option and partially satisfies a key differentiating detail of the 3-point option.",
  "Assign 1.5 only when the answer fully satisfies the 1-point option and partially satisfies a key differentiating detail of the 2-point option.",
  "Assign 0.5 only when the answer is relevant and experimentally reasonable but does not clearly match any positive rubric option.",
  "Do not assign half points merely because the answer is verbose, partially correct in a vague way, or scientifically plausible outside this scenario."
],

"Step 5 - Split scoring, only when explicitly needed": [
  "If split scoring is required, score each explicit sub-question using Steps 2-4.",
  "If one explicit sub-question is not answered, assign 0 to that sub-question.",
  "The final score is the mean of the explicit sub-question scores, rounded to the nearest 0.5.",
  "If split scoring is not required, output the single holistic score."
],

"Step 6 - Rationale requirements": [
  "The reason must state:",
  "1. Whether holistic scoring or split scoring was used, and why.",
  "2. The first substantive claim/procedure that was scored.",
  "3. The matched rubric option or score level.",
  "4. Why the answer did or did not meet the higher-score differentiating details.",
  "5. Whether any extra information was ignored, contradicted the scored point, or introduced safety risk."
]
},

"Output requirements:": [
  "1. Output exactly in this format:",
  "##JSON Output:",
  "{\"score\":2,\"reason\":\"...\"}",
  "2. The line after \"##JSON Output:\" must be one single valid JSON object.",
  "3. Do not output markdown code fences.",
  "4. Do not output extra text before or after the JSON object."
],

"final_instruction": "The JSON object must contain exactly two fields: \"score\" (float) and \"reason\" (string), in that order."
}
\end{lstlisting}

\subsection{Normalized-score calculation and judge aggregation}
Normalized scores were calculated from
question-level rubric scores as defined in
Section~3.4. For full-benchmark reporting,
judge-specific scores were first calculated
for each fine-grained subcategory and the
Materials Science category.
Major-category scores were obtained as
question-count-weighted averages of the
corresponding subcategory scores, and overall
scores were aggregated across the five major
categories using their question counts.
Judge-specific scores were rounded to one
decimal place at each aggregation level.
The final reported score was the arithmetic
mean of the two judge-specific scores, rounded
to one decimal place. Scores for the controlled
100-question subset were calculated directly
over the 100 questions.

\section{Expert Validation of Automated Scoring}
\label{app:Validation}


\begin{table*}[t]
\centering
\begingroup
\small
\setlength{\tabcolsep}{6.5pt}
\renewcommand{\arraystretch}{1.22}

\begin{tabular}{@{}lccccc@{}}
\toprule
\parbox[c][2.5\baselineskip][c]{3.0cm}{\raggedright\textbf{Scoring source}} &
\parbox[c][2.5\baselineskip][c]{1.8cm}{\centering\textbf{Preparation}} &
\parbox[c][2.5\baselineskip][c]{1.7cm}{\centering\textbf{Monitoring}} &
\parbox[c][2.5\baselineskip][c]{1.8cm}{\centering\textbf{Postprocess}} &
\parbox[c][2.5\baselineskip][c]{2.5cm}{\centering\textbf{Characterization}} &
\parbox[c][2.5\baselineskip][c]{1.9cm}{\centering\textbf{Material Science}} \\[1pt]
\midrule

GPT-5.5 judge
& 48.72\% & 63.89\% & 40.20\% & 47.62\% & 59.26\% \\

Gemini 3.1 Pro judge
& 50.00\% & 63.89\% & 40.39\% & 48.21\% & 60.19\% \\

\addlinespace[2pt]
Automated mean
& 49.36\% & 63.89\% & 40.29\% & 47.92\% & 59.72\% \\

\addlinespace[2pt]
Expert rater 1
& 48.72\% & 61.11\% & 40.00\% & 48.21\% & 58.33\% \\

Expert rater 2
& 47.44\% & 63.89\% & 40.20\% & 48.81\% & 59.26\% \\

Expert mean
& 48.08\% & 62.50\% & 40.10\% & 48.51\% & 58.80\% \\

\addlinespace[2pt]
Automated $-$ expert
& +1.28 pp & +1.39 pp & +0.19 pp & $-$0.59 pp & +0.92 pp \\

\bottomrule
\end{tabular}

\endgroup
\caption{Automated and expert scores by major category.}
\label{tab:si-8}
\end{table*}

Automated scoring was validated using responses generated by Claude Opus 4.7.
Questions were stratified across the five major categories-Experiment Preparation,
Experiment Monitor, Experiment Postprocess, Experiment Characterization, and
Materials Science-and approximately 20\% of the questions in each major category
were randomly sampled. Within each of the four organic-synthesis major categories,
sampling preserved as closely as possible the proportional distribution of the
corresponding fine-grained subcategories. Materials Science was sampled directly
because it was not further subdivided into fine-grained subcategories.
Two experts with relevant experimental experience independently rescored all sampled responses. The experts received the same scoring prompt used for the automated judges. During scoring, neither expert was provided with the automated-judge scores or the score assigned by the other expert.
All scores were normalized to percentages relative to the maximum score of 3. The mean automated score was calculated as the arithmetic mean of the scores from the GPT-5.5 and Gemini 3.1 Pro judges. The mean expert score was calculated as the arithmetic mean of the two independent expert ratings. Category-level differences were calculated as the mean automated score minus the mean expert score and are reported in percentage points (Table~\ref{tab:si-8}). Across the five major categories, the difference between mean automated and mean expert scores ranged from \ensuremath{-}0.59 to +1.39 percentage points. The dual-judge mean achieved a mean absolute error
(MAE) of 0.165 points on the 0-3 scale relative
to the two-expert mean, with 98.0\% of responses
differing by no more than 0.5 points
(quadratic-weighted $\kappa=0.9161$).
The individual MAEs were 0.185 and 0.195 for
GPT-5.5 and Gemini 3.1 Pro, respectively.
The two experts had an inter-rater MAE of
0.330 points ($\kappa=0.7515$).

\section{Model Performance on subcategories}

To provide the complete fine-grained results underlying the major-category performance reported in the main text, model scores were further summarized for each organic-synthesis subcategory and for the materials-science category, as shown in Table~\ref{tab:si-13} to \ref{tab:si-12}. The tables below report the two judge-specific normalized scores for each evaluated model within each fine-grained task, together with the number of questions in each subcategory.

The organic-synthesis benchmark was divided into Experiment Preparation, Experiment Monitoring, Experiment Postprocess, and Experiment Characterization, with the corresponding fine-grained subcategories shown below. Materials Science was retained as a separate major category and was not further subdivided in this analysis.

In each table entry, the value to the left of the slash represents the GPT-5.5 judge score, and the value to the right represents the Gemini 3.1 Pro judge score.

\newcommand{\SIpair}[2]{\mbox{#1\%\,/\,#2\%}}

\begin{table}[H]
\centering
\begingroup
\footnotesize
\setlength{\tabcolsep}{3pt}
\renewcommand{\arraystretch}{1.15}
\textbf{Materials Science ($n=92$)}\par\vspace{4pt}
\begin{tabularx}{\columnwidth}{@{}
>{\raggedright\arraybackslash}X
>{\centering\arraybackslash}p{0.39\columnwidth}@{}}
\toprule
\textbf{Model} & \textbf{Judge scores} \\
\midrule
Claude Opus 4.7       & \SIpair{58.2}{61.4} \\
Gemini 3.1 Pro        & \SIpair{51.1}{55.8} \\
Gemini 3.1 Flash-Lite & \SIpair{46.0}{49.1} \\
GPT-5.5               & \SIpair{52.4}{53.3} \\
Grok 4.3              & \SIpair{52.0}{55.3} \\
Qwen3.6-Plus          & \SIpair{50.2}{55.3} \\
MiniMax-M3            & \SIpair{54.0}{57.1} \\
Intern-S1-Pro         & \SIpair{41.3}{42.0} \\
\bottomrule
\end{tabularx}
\endgroup
\caption{Model performance for materials-science tasks.}
\label{tab:si-13}
\vspace{-10pt}
\end{table}

\begin{table}[H]
\centering
\begingroup
\footnotesize
\setlength{\tabcolsep}{2pt}
\renewcommand{\arraystretch}{1.15}
\textbf{Experiment Preparation ($n=67$)}\par\vspace{4pt}
\begin{tabularx}{\columnwidth}{@{}
>{\raggedright\arraybackslash}p{0.32\columnwidth}
>{\centering\arraybackslash}X
>{\centering\arraybackslash}X@{}}
\toprule
\textbf{Model} &
\shortstack[c]{\textbf{Experiment\_}\\\textbf{plan ($n=46$)}} &
\shortstack[c]{\textbf{Experiment\_}\\\textbf{Safety ($n=21$)}} \\
\midrule
Claude Opus 4.7        & \SIpair{54.7}{54.7} & \SIpair{33.3}{40.5} \\
Gemini 3.1 Pro         & \SIpair{41.7}{43.1} & \SIpair{29.4}{27.0} \\
Gemini 3.1 Flash-Lite  & \SIpair{44.9}{47.5} & \SIpair{31.0}{28.6} \\
GPT-5.5                & \SIpair{50.4}{49.6} & \SIpair{42.9}{50.8} \\
Grok 4.3               & \SIpair{37.3}{39.1} & \SIpair{27.8}{33.3} \\
Qwen3.6-Plus           & \SIpair{40.2}{41.7} & \SIpair{26.2}{32.5} \\
MiniMax-M3             & \SIpair{49.3}{48.6} & \SIpair{42.9}{44.4} \\
Intern-S1-Pro          & \SIpair{25.0}{23.6} & \SIpair{24.6}{23.0} \\
\bottomrule
\end{tabularx}
\endgroup
\caption{Model performance for experiment-preparation subcategories.}
\label{tab:si-9}
\vspace{-10pt}
\end{table}

\begin{table}[H]
\centering
\begingroup
\footnotesize
\setlength{\tabcolsep}{2pt}
\renewcommand{\arraystretch}{1.15}
\textbf{Experiment Monitoring ($n=30$)}\par\vspace{4pt}
\begin{tabularx}{\columnwidth}{@{}
>{\raggedright\arraybackslash}p{0.32\columnwidth}
>{\centering\arraybackslash}X
>{\centering\arraybackslash}X@{}}
\toprule
\textbf{Model} &
\shortstack[c]{\textbf{Experiment\_}\\\textbf{Monitor ($n=16$)}} &
\shortstack[c]{\textbf{Experiment\_}\\\textbf{notes ($n=14$)}} \\
\midrule
Claude Opus 4.7        & \SIpair{44.8}{47.9} & \SIpair{85.7}{86.9} \\
Gemini 3.1 Pro         & \SIpair{32.3}{35.4} & \SIpair{64.3}{64.3} \\
Gemini 3.1 Flash-Lite  & \SIpair{31.3}{31.3} & \SIpair{52.4}{53.6} \\
GPT-5.5                & \SIpair{43.8}{41.7} & \SIpair{61.9}{65.5} \\
Grok 4.3               & \SIpair{33.3}{41.7} & \SIpair{61.9}{64.3} \\
Qwen3.6-Plus           & \SIpair{36.5}{36.5} & \SIpair{76.2}{75.0} \\
MiniMax-M3             & \SIpair{44.8}{40.6} & \SIpair{65.5}{71.4} \\
Intern-S1-Pro          & \SIpair{21.9}{26.0} & \SIpair{58.3}{58.3} \\
\bottomrule
\end{tabularx}
\endgroup
\caption{Model performance for experiment-monitoring subcategories.}
\label{tab:si-10}
\end{table}

\begin{table*}[!tp]
\centering
\begingroup
\footnotesize
\setlength{\tabcolsep}{3pt}
\renewcommand{\arraystretch}{1.14}
\textbf{Postprocess ($n=425$)}\par\vspace{5pt}
\begin{tabularx}{\textwidth}{@{}
>{\raggedright\arraybackslash}p{0.19\textwidth}
*{4}{>{\centering\arraybackslash}X}@{}}
\toprule
\textbf{Model} &
\shortstack[c]{\textbf{Extraction}\\($n=38$)} &
\shortstack[c]{\textbf{Evaporation}\\($n=46$)} &
\shortstack[c]{\textbf{Column}\\($n=75$)} &
\shortstack[c]{\textbf{Recrystallization}\\($n=16$)} \\
\midrule
Claude Opus 4.7       & \SIpair{48.2}{49.6} & \SIpair{48.9}{51.4} & \SIpair{44.9}{46.2} & \SIpair{32.3}{33.3} \\
Gemini 3.1 Pro        & \SIpair{46.5}{44.3} & \SIpair{39.1}{40.6} & \SIpair{39.1}{43.1} & \SIpair{31.3}{36.5} \\
Gemini 3.1 Flash-Lite & \SIpair{43.0}{42.1} & \SIpair{37.0}{37.0} & \SIpair{29.6}{32.7} & \SIpair{28.1}{35.4} \\
GPT-5.5               & \SIpair{43.4}{46.5} & \SIpair{50.7}{50.4} & \SIpair{46.9}{46.7} & \SIpair{34.4}{38.5} \\
Grok 4.3              & \SIpair{46.1}{44.7} & \SIpair{36.2}{36.6} & \SIpair{40.4}{41.3} & \SIpair{33.3}{36.5} \\
Qwen3.6-Plus          & \SIpair{46.5}{46.9} & \SIpair{40.2}{40.2} & \SIpair{36.7}{42.2} & \SIpair{22.9}{30.2} \\
MiniMax-M3            & \SIpair{43.0}{44.7} & \SIpair{39.9}{39.9} & \SIpair{34.4}{36.0} & \SIpair{34.4}{32.3} \\
Intern-S1-Pro         & \SIpair{35.1}{36.8} & \SIpair{27.9}{23.2} & \SIpair{26.9}{28.2} & \SIpair{26.0}{29.2} \\
\bottomrule
\end{tabularx}
\par\vspace{8pt}
\begin{tabularx}{\textwidth}{@{}
>{\raggedright\arraybackslash}p{0.19\textwidth}
*{3}{>{\centering\arraybackslash}X}@{}}
\toprule
\textbf{Model} &
\shortstack[c]{\textbf{Trituration}\\($n=25$)} &
\shortstack[c]{\textbf{Lyophilization}\\($n=24$)} &
\shortstack[c]{\textbf{TLC}\\($n=201$)} \\
\midrule
Claude Opus 4.7       & \SIpair{36.7}{38.0} & \SIpair{44.4}{43.8} & \SIpair{35.0}{34.9} \\
Gemini 3.1 Pro        & \SIpair{24.0}{24.0} & \SIpair{42.4}{45.1} & \SIpair{35.0}{35.7} \\
Gemini 3.1 Flash-Lite & \SIpair{22.7}{18.0} & \SIpair{39.6}{39.6} & \SIpair{32.5}{31.3} \\
GPT-5.5               & \SIpair{33.3}{33.3} & \SIpair{46.5}{47.9} & \SIpair{32.1}{32.3} \\
Grok 4.3              & \SIpair{34.0}{33.3} & \SIpair{38.2}{38.9} & \SIpair{27.8}{26.5} \\
Qwen3.6-Plus          & \SIpair{26.7}{28.0} & \SIpair{36.1}{36.8} & \SIpair{36.0}{36.0} \\
MiniMax-M3            & \SIpair{22.0}{15.3} & \SIpair{46.5}{47.2} & \SIpair{30.1}{28.3} \\
Intern-S1-Pro         & \SIpair{18.0}{17.3} & \SIpair{36.1}{36.1} & \SIpair{23.4}{21.6} \\
\bottomrule
\end{tabularx}
\endgroup
\caption{Model performance for postprocessing subcategories.}
\label{tab:si-11}
\end{table*}

\begin{table*}[!tp]
\centering
\begingroup
\footnotesize
\setlength{\tabcolsep}{2.5pt}
\renewcommand{\arraystretch}{1.14}
\textbf{Experiment Characterization ($n=142$)}\par\vspace{5pt}
\begin{tabularx}{\textwidth}{@{}
>{\raggedright\arraybackslash}p{0.19\textwidth}
*{5}{>{\centering\arraybackslash}X}@{}}
\toprule
\textbf{Model} &
\shortstack[c]{\textbf{NMR}\\($n=42$)} &
\shortstack[c]{\textbf{LCMS}\\($n=48$)} &
\shortstack[c]{\textbf{Meltingpoint}\\($n=8$)} &
\shortstack[c]{\textbf{Crystal}\\($n=29$)} &
\shortstack[c]{\textbf{GCMS}\\($n=15$)} \\
\midrule
Claude Opus 4.7       & \SIpair{41.3}{41.7} & \SIpair{57.6}{59.0} & \SIpair{8.3}{6.3}   & \SIpair{55.7}{55.7} & \SIpair{34.4}{33.3} \\
Gemini 3.1 Pro        & \SIpair{47.2}{48.8} & \SIpair{54.2}{56.6} & \SIpair{18.8}{25.0} & \SIpair{49.4}{54.0} & \SIpair{31.1}{31.1} \\
Gemini 3.1 Flash-Lite & \SIpair{38.1}{42.1} & \SIpair{36.8}{37.2} & \SIpair{16.7}{14.6} & \SIpair{40.8}{45.4} & \SIpair{31.1}{40.0} \\
GPT-5.5               & \SIpair{44.4}{48.0} & \SIpair{54.2}{53.8} & \SIpair{16.7}{16.7} & \SIpair{47.1}{50.6} & \SIpair{38.9}{42.2} \\
Grok 4.3              & \SIpair{31.0}{31.3} & \SIpair{45.5}{50.0} & \SIpair{12.5}{14.6} & \SIpair{42.0}{40.8} & \SIpair{24.4}{28.9} \\
Qwen3.6-Plus          & \SIpair{42.1}{44.0} & \SIpair{55.6}{54.9} & \SIpair{22.9}{20.8} & \SIpair{36.8}{37.9} & \SIpair{42.2}{43.3} \\
MiniMax-M3            & \SIpair{36.9}{37.3} & \SIpair{46.9}{49.3} & \SIpair{16.7}{16.7} & \SIpair{36.8}{36.8} & \SIpair{52.2}{52.2} \\
Intern-S1-Pro         & \SIpair{31.0}{29.4} & \SIpair{29.5}{31.3} & \SIpair{10.4}{8.3}  & \SIpair{32.8}{29.3} & \SIpair{35.6}{31.1} \\
\bottomrule
\end{tabularx}
\endgroup
\caption{Model performance for characterization subcategories.}
\label{tab:si-12}
\end{table*}



\FloatBarrier

\begin{table*}[t]
\centering
\small
\setlength{\tabcolsep}{5pt}
\renewcommand{\arraystretch}{1.15}

\begin{tabularx}{\textwidth}{@{}
>{\raggedright\arraybackslash}p{0.19\textwidth}
>{\centering\arraybackslash}p{0.14\textwidth}
>{\centering\arraybackslash}p{0.09\textwidth}
>{\raggedright\arraybackslash}X
@{}}

\toprule

\textbf{Major category}
&
\makecell[c]{\textbf{Full benchmark}\\\textbf{$N$}}
&
\makecell[c]{\textbf{Subset}\\\textbf{$n$}}
&
\makecell[l]{\textbf{Fine-grained allocation}\\
\textbf{Full $N \rightarrow$ subset $n$}}
\\

\midrule

Preparation
& 67
& 10
& Experiment\_plan: 46 $\rightarrow$ 7;
  Experiment\_Safety: 21 $\rightarrow$ 3
\\[3pt]

Monitoring
& 30
& 5
& Experiment\_Monitor: 16 $\rightarrow$ 3;
  Experiment\_notes: 14 $\rightarrow$ 2
\\[3pt]

Postprocessing
& 425
& 64
& Extraction: 38 $\rightarrow$ 6;
  Evaporation: 46 $\rightarrow$ 7;
  Column: 75 $\rightarrow$ 11;
  TLC: 201 $\rightarrow$ 30;
  Recrystallization: 16 $\rightarrow$ 2;
  Trituration: 25 $\rightarrow$ 4;
  Lyophilization: 24 $\rightarrow$ 4
\\[3pt]

Characterization
& 142
& 21
& NMR: 42 $\rightarrow$ 6;
  LCMS: 48 $\rightarrow$ 7;
  Meltingpoint: 8 $\rightarrow$ 1;
\mbox{Crystal: 29 $\rightarrow$ 5};
\mbox{GCMS: 15 $\rightarrow$ 2}
\\

\midrule

\textbf{Total}
& \textbf{664}
& \textbf{100}
& \textbf{16 subcategories}
\\

\bottomrule

\end{tabularx}
\caption{Composition of the controlled 100-question subset.}
\label{tab:si-14}
\end{table*}

\section{100-question subset and human experiment design}
\label{app:100-question subset}

We constructed a subset of 100 questions from the 664 organic-synthesis questions in the benchmark. The subset used fixed, approximately proportional quotas at two taxonomy levels. Within each major category, its quota was distributed across the constituent fine-grained subcategories approximately in proportion to their prevalence in the corresponding source category (Table~\ref{tab:si-14}).

\subsection{Human evaluation procedure}
Ten participants were included in the human evaluation, comprising five undergraduate participants and five PhD participants. Participants in both groups had hands-on laboratory experience relevant to the experimental situations represented in the subset.
Each participant independently completed all 100 questions in the controlled organic-synthesis subset. Participants were provided with the same question text, experimental context and associated visual evidence used for model evaluation and were instructed to answer the specific experimental judgment requested by each question.

Responses were expected to be direct and concise while addressing the required judgment. When a question contained multiple explicitly stated parts, participants were asked to address each part. Participants completed the questions independently, without discussion with other participants.
Participants were instructed to answer using only the information supplied with each question together with their own laboratory knowledge and experience. External references, AI-assisted tools and other external resources were not permitted during the evaluation.

\subsection{Human-response scoring}
Human answers were matched to the subset by question ID. The controlled-subset JSON served as the sole source of question order, category labels, original question text and item-specific scoring rubrics; the participant spreadsheets supplied only the human responses. This ensured that each human response was associated with exactly the same question-specific rubric used for evaluating model responses.

Each response was scored independently by two automated judges, GPT-5.5 and Gemini 3.1 Pro, using the same scoring program and evaluation prompt applied to model responses. Each judge received the original question text, the candidate response and the complete item-specific A-D rubric. Participant identity and group membership were not included in the judge prompt. Images were not separately provided to the scoring judges, consistent with the model-response scoring procedure.
The rubric was treated as authoritative. Each judge assigned a score from 0 to 3, with half-point scores permitted under the predefined scoring rules.
For participant \(p\) and judge \(j\), the judge-specific normalized score was

\begin{equation*}
S_{p,j} = 100\times
\frac{\sum_{i\in V_{p,j}} s_{p,i,j}}{3\lvert V_{p,j}\rvert},
\end{equation*}

where \(s_{p,i,j}\) is the score assigned to question \(i\), and \(V_{p,j}\) is the set of questions with valid judge scores. The reported two-judge score was the arithmetic mean

\begin{equation*}
S_p = \frac{S_{p,\mathrm{GPT\text{-}5.5}} + S_{p,\mathrm{Gemini}}}{2}.
\end{equation*}

All final human-response files contained valid scores for all 100 questions from both judges; therefore, \(|V_{p,j}|=100\) for every participant. Group-level results were calculated across five participants per training group and are reported as the mean and sample standard deviation.


Individual participant performance on the controlled 100-question organic-synthesis subset is reported below (Table~\ref{tab:si-15}). Each participant's responses were independently scored by GPT-5.5 and Gemini 3.1 Pro using the same automated scoring procedure applied to model responses. The mean score for each participant was calculated from the two judge-specific normalized scores. Group-level variability was calculated as the sample standard deviation across the five participants in each group. The undergraduate group achieved a mean normalized score of 31.0\% (sample SD = 2.7 percentage points), whereas the PhD group achieved a mean normalized score of 45.9\% (sample SD = 3.0 percentage points).

\begin{table*}[t]

\centering
{\small
\setlength{\tabcolsep}{2.1pt}
\renewcommand{\arraystretch}{1.16}
\begin{tabularx}{\textwidth}{@{}>{\raggedright\arraybackslash\hsize=1.0000\hsize}X>{\raggedright\arraybackslash\hsize=1.4000\hsize}X>{\centering\arraybackslash\hsize=1.0000\hsize}X>{\centering\arraybackslash\hsize=1.0500\hsize}X>{\centering\arraybackslash\hsize=0.8500\hsize}X>{\centering\arraybackslash\hsize=0.7000\hsize}X@{}}
\toprule
\textbf{Participant ID} & \textbf{Group} & \textbf{GPT-5.5 judge} & \textbf{Gemini 3.1 Pro judge} & \textbf{Mean score} & \textbf{Sample SD} \\
\midrule
U01 & Undergraduate & 26.7\% & 28.2\% & 27.4\% & --- \\
U02 & Undergraduate & 29.0\% & 31.3\% & 30.2\% & --- \\
U03 & Undergraduate & 33.8\% & 35.0\% & 34.4\% & --- \\
U04 & Undergraduate & 29.8\% & 30.8\% & 30.3\% & --- \\
U05 & Undergraduate & 32.0\% & 33.3\% & 32.7\% & --- \\
Group mean & Undergraduate, n = 5 & 30.3\% & 31.7\% & 31.0\% & 2.7\% \\
P01 & PhD & 45.3\% & 49.3\% & 47.3\% & --- \\
P02 & PhD & 44.7\% & 48.5\% & 46.6\% & --- \\
P03 & PhD & 39.5\% & 42.0\% & 40.8\% & --- \\
P04 & PhD & 46.2\% & 50.7\% & 48.4\% & --- \\
P05 & PhD & 44.7\% & 48.3\% & 46.5\% & --- \\
Group mean & PhD, n = 5 & 44.1\% & 47.8\% & 45.9\% & 3.0\% \\
\bottomrule
\end{tabularx}
}

\caption{Individual human participant scores.}\label{tab:si-15}
\end{table*}


\section{Mismatched-Image Ablation}
\label{app:mismatch}
A mismatched-image ablation was performed to quantify model sensitivity to image-question alignment on the 100-question organic-synthesis subset (Table~\ref{tab:si-16}). The intervention disrupted the correspondence between the visual evidence and the question while retaining question identity, question text, text-encoded experimental context, answer-generation instructions, question-specific scoring rubric and evaluation pipeline.
For each question in the matched-image subset, the original image or images were replaced with mismatched images sampled from other benchmark items within the same fine-grained subcategory. The target question's original image or images were excluded from the replacement set. The number of images per question was kept unchanged, while the question text and all other inputs were preserved.
The matched and mismatched datasets contained the same 100 unique question IDs in the same order. Direct question-level comparison confirmed that the category, question text, question-specific 0--3 rubric, reference answer and original subset metadata were identical; only the image array changed, with additional provenance fields added to the mismatched dataset.
Before inference, literal image filenames in the question were replaced with neutral labels such as ``Figure 1''. For mismatched questions, this sanitization used the original-image references rather than the distractor filenames, preventing the replacement filename from revealing the intervention.

\begin{table*}[t]

\centering
{\small
\setlength{\tabcolsep}{1.3pt}
\renewcommand{\arraystretch}{1.16}
\begin{tabularx}{\textwidth}{
@{}
>{\raggedright\arraybackslash\hsize=1.1000\hsize}X
>{\centering\arraybackslash\hsize=1.2800\hsize}X
>{\centering\arraybackslash\hsize=0.7200\hsize}X
>{\centering\arraybackslash\hsize=1.3000\hsize}X
>{\centering\arraybackslash\hsize=0.9000\hsize}X
>{\centering\arraybackslash\hsize=0.7000\hsize}X
@{}}
\toprule
\textbf{Model} &
\textbf{Matched scores} &
\textbf{Matched mean} &
\textbf{Mismatched scores} &
\textbf{Mismatched mean} &
\textbf{Score drop} \\
\midrule
Claude Opus 4.7 & 43.2\% / 44.2\% & 43.7\% & 25.0\% / 20.3\% & 22.7\% & 21.0 pp \\
Gemini 3.1 Pro & 37.8\% / 40.2\% & 39.0\% & 17.7\% / 16.7\% & 17.2\% & 21.8 pp \\
Gemini 3.1 Flash-Lite & 34.7\% / 36.0\% & 35.4\% & 22.5\% / 18.7\% & 20.6\% & 14.8 pp \\
GPT-5.5 & 41.7\% / 45.2\% & 43.5\% & 21.0\% / 18.0\% & 19.5\% & 24.0 pp \\
Grok 4.3 & 35.5\% / 33.5\% & 34.5\% & 22.5\% / 20.7\% & 21.6\% & 12.9 pp \\
Qwen3.6-Plus & 38.2\% / 39.7\% & 39.0\% & 20.5\% / 18.3\% & 19.4\% & 19.6 pp \\
MiniMax-M3 & 37.5\% / 38.0\% & 37.8\% & 21.0\% / 18.5\% & 19.8\% & 18.0 pp \\
Intern-S1-Pro & 27.5\% / 26.3\% & 26.9\% & 13.5\% / 14.2\% & 13.9\% & 13.0 pp \\
Cross-model mean & 37.0\% / 37.9\% & 37.5\% & 20.5\% / 18.2\% & 19.3\% & 18.2 pp \\
\bottomrule
\end{tabularx}
}
\caption{The cross-model score drop is the difference
between the separately rounded condition means,
rather than the mean of model-wise reductions.}
\label{tab:si-16}
\end{table*}

\begin{figure*}[!t]
\centering
\includegraphics[
    width=0.85\textwidth
]{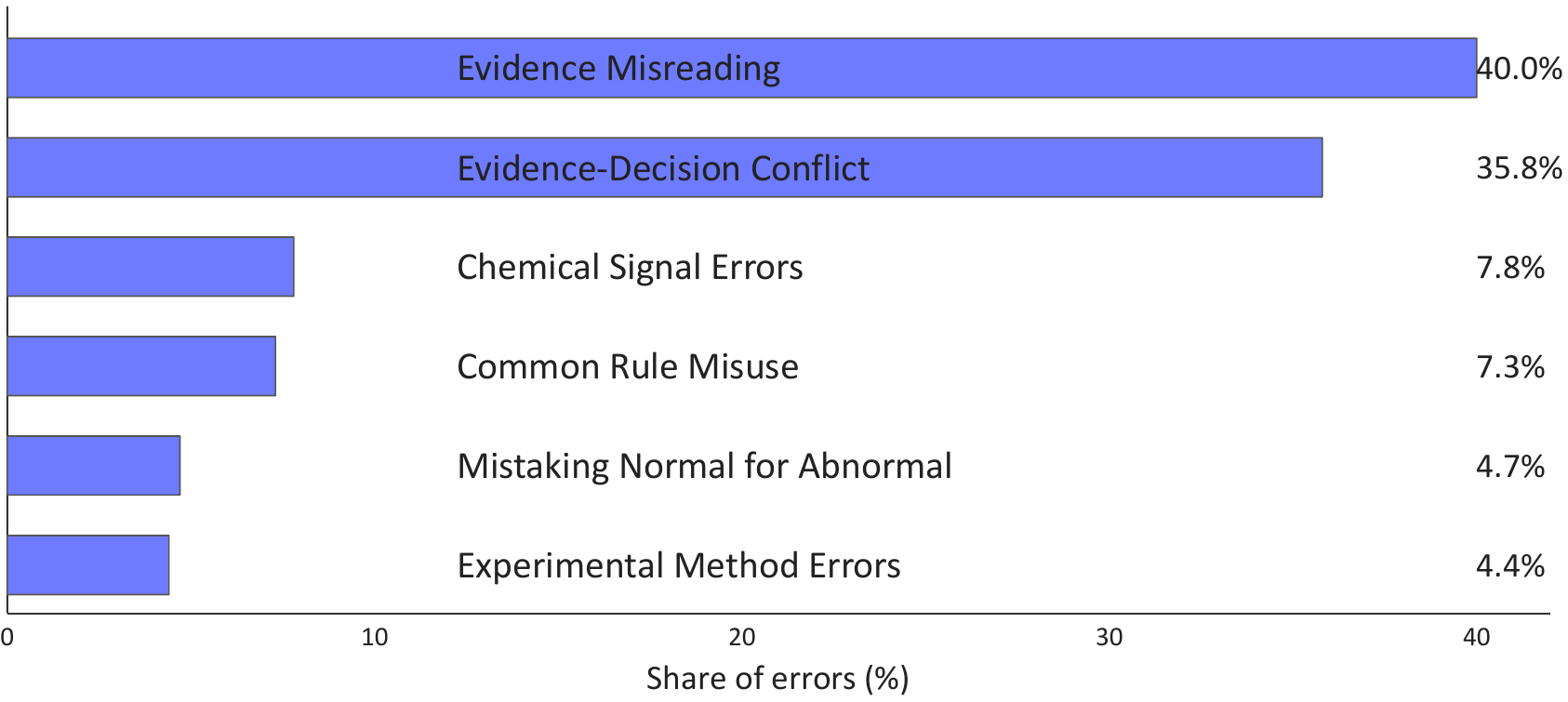}
\caption{Model-balanced shares of assigned error labels across Claude Opus 4.7, GPT-5.5, and Gemini 3.1 Pro on the controlled organic-synthesis subset. Percentages sum to 100\%.}
\label{fig:wrongreason}
\end{figure*}

\begin{table*}[t]
\centering
\small
\setlength{\tabcolsep}{4pt}
\renewcommand{\arraystretch}{1.12}

\begin{tabularx}{\textwidth}{@{}
>{\raggedright\arraybackslash}p{0.20\textwidth}
>{\raggedright\arraybackslash}X
>{\centering\arraybackslash}p{0.09\textwidth}
>{\centering\arraybackslash}p{0.09\textwidth}
>{\centering\arraybackslash}p{0.09\textwidth}
@{}}

\toprule

\textbf{Error type} &
\textbf{Definition} &
\shortstack[c]{\textbf{GPT-}\\\textbf{5.5}} &
\shortstack[c]{\textbf{Gemini}\\\textbf{3.1 Pro}} &
\shortstack[c]{\textbf{Claude}\\\textbf{Opus 4.7}} \\

\midrule

Evidence Misreading &
Misreading or omission of decision-relevant experimental evidence. &
38 & 31 & 34 \\

\addlinespace[3pt]

Evidence--Decision Conflict &
A diagnosis or action inconsistent with the experimental evidence or rubric. &
33 & 29 & 30 \\

\addlinespace[3pt]

Chemical Signal Errors &
Incorrect interpretation of chemical signals, peaks, species, or quantitative results. &
6 & 5 & 9 \\

\addlinespace[3pt]

Common Rule Misuse &
Applying generic laboratory rules in place of case-specific evidence. &
9 & 6 & 4 \\

\addlinespace[3pt]

Mistaking Normal for Abnormal &
Treating a normal or transient state as abnormal and recommending unnecessary intervention. &
3 & 5 & 4 \\

\addlinespace[3pt]

Experimental Method Errors &
Independent errors in experimental steps, conditions, methods, or execution. &
3 & 5 & 3 \\

\midrule

\textbf{Total assignments} &
&
\textbf{92} &
\textbf{81} &
\textbf{84} \\

\bottomrule

\end{tabularx}

\caption{Definitions and raw counts of expert-assigned error labels across the three models.}
\label{tab:si-18}

\end{table*}

\subsection{Ablation results}

For model (m), condition (c), and judge (j), the judge-specific normalized score was calculated over the 100-question subset as

\begin{equation*}
NS_{j,m,c} =
\frac{\sum_{i=1}^{100} s_{i,j,m,c}}{3\times 100}
\times 100\%,
\end{equation*}

where \(s_{i,j,m,c}\) is the question-level score assigned by judge (j).

The normalized score was calculated separately for GPT-5.5 and Gemini 3.1 Pro. Each judge-specific normalized score was first rounded to one decimal place. The final score for each model under each image condition was then calculated as the arithmetic mean of the two rounded judge-specific normalized scores:

\begin{equation*}
S_{m,c} =
\frac{
NS_{\mathrm{GPT\text{-}5.5,m,c}} +
NS_{\mathrm{Gemini,3.1,Pro},m,c}
}{2}.
\end{equation*}
Cross-model means were calculated as the arithmetic mean of the eight displayed final model scores under each condition and rounded to one decimal place.
For each model, the score reduction was calculated as the difference between the original and perturbed scores and is reported in percentage points.

\begin{equation*}
\Delta_m = S_{m,\mathrm{matched}} - S_{m,\mathrm{mismatched}},
\end{equation*}


Values are displayed to one decimal place. Before final display rounding, the matched-image aggregate scores of Gemini 3.1 Pro and Qwen3.6-Plus were 39.00\% and 38.95\%, respectively.
All eight matched and mismatched evaluation batches contained 100 model responses and 100 valid scores from each judge, with no remaining answer-generation errors, scoring errors or missing scores. Cross-model means were calculated separately
for the matched- and mismatched-image conditions
and rounded to one decimal place. The reported
18.2-percentage-point reduction is defined as
the difference between these rounded means
(37.5\% - 19.3\% = 18.2 percentage points).
This differs slightly from the mean of the eight
model-wise reductions (18.1 percentage points)
because of the order of rounding and aggregation.

\section{Expert Error Analysis}
\label{app:error}

Two experts independently reviewed all non-full-score responses from Claude Opus 4.7, GPT-5.5, and Gemini 3.1 Pro on the 100-question organic-synthesis subset, using the corresponding questions, visual evidence, reference answers, and automated scoring rationales. Each expert identified errors in evidence interpretation, causal reasoning, or experimental decisions. Their independent annotations were compared and consolidated into a shared six-category taxonomy, with the annotation rules finalized during this process.

Of the 266 non-full-score responses (83 from Claude Opus 4.7, 89 from GPT-5.5, and 94 from Gemini 3.1 Pro), 159 received at least one error label, yielding 257 label assignments. The taxonomy was non-exhaustive; responses without a clearly identifiable error mechanism remained unlabeled. Table~\ref{tab:si-18} summarizes the error types and their raw counts.





\paragraph{Annotation and aggregation rules.}
Evidence Misreading, Common Rule Misuse, and Evidence--Decision Conflict could co-occur when independently supported. The remaining three error types were mutually exclusive with one another and with these evidence-to-decision categories. Errors in experimental methods arising directly from an incorrect diagnosis were classified as Evidence-Decision Conflict, while chemical signal misassignments were classified as Chemical Signal Errors.

For each error type, its share of label assignments was calculated within each model and then averaged equally across the three models, rather than pooling all 257 assignments. Percentages were rounded to one decimal place using the largest-remainder method to sum to 100.0\%. These values describe the model-balanced composition of assigned error labels, not response-level error rates.

\end{document}